\documentclass[preprint,12pt]{elsarticle}
\usepackage{amssymb}
\newcommand{\ignore}[1]{}
\usepackage{graphicx}
\usepackage{hyperref}
\usepackage{url}
\usepackage{amsmath,amssymb} 
\usepackage{algorithmic}
\usepackage{algorithm}
\usepackage{multirow}
\usepackage{color}
\usepackage{setspace}
\usepackage{CJK}

\journal{Pattern Recognition}

\begin{document}

\begin{frontmatter}

\title{Adaptive Compressive Tracking via Online Vector Boosting Feature Selection}
\author{Qingshan Liu}
\author{Jing Yang}
\author{Kaihua Zhang\corref{cor1}}
\ead{zhkhua@gmail.com}
\cortext[cor1]{Corresponding author. Phone number: 086-13851581017}
\author[]{Yi Wu}
\address{Jiangsu Key Laboratory of Big Data Analysis Technology, Nanjing University of Information Science and Technology, Nanjing, China}

\begin{abstract}
Recently, the compressive tracking (CT) method~\cite{zhang2012real} has attracted much attention due to its high efficiency, but it cannot well deal with the large scale target appearance variations due to its data-independent random projection matrix that results in less discriminative features. To address this issue, in this paper we propose an adaptive CT approach, which selects the most discriminative features to design an effective appearance model. Our method significantly improves CT in three aspects: Firstly, the most discriminative features are selected via an online vector boosting method. Secondly, the object representation is updated in an effective online manner, which preserves the stable features while filtering out the noisy ones. Finally, a simple and effective trajectory rectification approach is adopted that can make the estimated location more accurate. Extensive experiments on the CVPR2013 tracking benchmark demonstrate the superior performance of our algorithm compared over  state-of-the-art tracking algorithms.
\end{abstract}

\end{frontmatter}

\section{Introduction}

Object tracking is a fundamental problem in computer vision with numerous applications such as motion analysis, surveillance, autonomous robots,etc, and much process has been witnessed in recent years~\cite{li2013survey}. However, it remains a challenging task due to the factors like illumination changes, partial occlusion, deformation, as well as viewpoint variation, to name a few~\cite{yilmaz2006object}. To well handle these factors, an effective appearance model is of great importance, in which numerous design schemes have been proposed~\cite{grabner2006real,grabner2008semi,jia2012visual,zhang2012robust,kwon2010visual,kwon2011tracking,babenko2011robust,kalal2010pn,zhang2012real}, which can be categorized into either generative models or discriminative ones.

Generative models learn an appearance model with the object information, which is used to search for the object with the minimum reconstruction error within a certain region.
Adam et al.~\cite{adam2006robust} represent the target appearance with the intensity histograms of multiple fragments that can be efficiently computed by integral images. Ross et al.~\cite{ross2008incremental} present a tracking method that incrementally learns a low-dimensional subspace representation, which can effectively adapt to the target appearance changes.
Mei and Ling~\cite{mei2009robust} treat tracking as a sparse representation problem, in which the target location is determined by solving an $\ell_1$ minimization problem. Bao et al.~\cite{bao2012real} utilize the accelerated proximal gradient approach to efficiently solve the $\ell_1$ minimization problem for visual tracking. In~\cite{kwon2010visual}, Kwon and Lee propose a visual tracking decomposition method that combines multiple observation and motion models for robust visual tracking, which has been further extended to search for appropriate trackers by the Markov Chain Monte Carlo sampling method~\cite{kwon2011tracking}. Zhang et al.~\cite{zhang2012robust} formulate the tracking task as a multi-task sparse learning problem.
In~\cite{jia2012visual}, Jia et al. formulate the object appearance as sparse codings of  the local image patches with their spatial layout.
In~\cite{zhang2012robust}, Zhong et al. propose a collaborative tracking algorithm that combines a sparsity-based discriminative classifier and a sparsity-based generative model. Wang et al.~\cite{wang2013least} present a least soft-threshold squares algorithm that models the image noise with the Gaussian-Laplacian distribution.

Discriminative models learn a binary classifier to distinguish the target from its surrounding background.
Avidan~\cite{avidan2004support} first proposes to utilize a support vector machine classifier for visual tracking. In~\cite{collins2005online}, an online discriminative feature selection technique is proposed to extract the most discriminative features to separate the target from the background. Grabner et al.~\cite{grabner2006real} proposes an online boosting method to select features for visual tracking. Babenko et al.~\cite{babenko2011robust} formulate the tracking task as a multiple instance learning (MIL) problem, and propose an online MIL boosting method that selects features to design an appearance model. Zhang and Song~\cite{zhang2013real} further extend the MIL tracking method by considering the sample importance. Kalal et al.~\cite{kalal2010pn} integrate a re-detection module into tracking that can restart tracking after the target reappears when it is completely occluded or missing from the scene. Hare et al.~\cite{hare2011struck} exploit the constraints of the predicted outputs with a kernelized structured SVM classifier, which achieves favorable results on the CVPR2013 tracking benchmark~\cite{wu2013online}. Henriques et al.~\cite{henriques2012exploiting} propose a fast tracking algorithm which explores the circulant structure of the kernel matrix in the SVM classifier that can be efficiently computed by the fast Fourier transform algorithm. Zhang et al.~\cite{zhang2012real} propose a real-time compressive tracking (CT) algorithm that employs a very sparse random matrix to achieve a low-dimensional image representation. In~\cite{zhang2014fast} Zhang et al. further reduce the computational complexity of CT with a coarse-to-fine strategy. Song~\cite{song2014robust} improves the performance of CT by introducing informative feature selection strategy.

Recently, Wu et al.~\cite{wu2013online} release the CVPR2013 tracking benchmark, which contains $50$ challenging sequences ($\sim26000$ frames), most of which suffer large scale target appearance variations. Results of $29$ tracking algorithms are reported including most above mentioned tracking algorithms. Although CT is very efficient that runs over $60$ frames per second (FPS), its success rate of one-pass evaluation (OPE) is only $30.6\%$. We claim that the unfavorable performance of CT is due to its data-independent random projection matrix that can only yield fixed feature templates, which cannot adapt the large scale target appearance variations well. In this paper, we propose an adaptive CT method which selects the most discriminative patches to construct the feature templates via a novel online vector boosting method. Furthermore, we adopt a new model update mechanism that can preserve the stable features while avoiding the noisy ones, thereby effectively alleviating the drift problem caused by online model update. Finally, we propose a very simple trajectory rectification method that makes the finally estimated location more accurate.
Extensive evaluations on the CVPR2013 tracking benchmark~\cite{wu2013online} demonstrate the proposed algorithm performs favorably against state-of-art methods in terms of efficiency, accuracy and robustness, and especially the proposed algorithm outperforms CT by a large margin (the success rate of OPE of the proposed method is $50.4\%$ vs. $30.6\%$ for CT).

The key contributions of the proposed algorithm are summarized as follows:
\begin{itemize}
  \item  To the best of our knowledge, it is the first time to explore online vector boosting~\cite{huang2005vector} feature selection method for visual tracking, in which the selected features can well adapt to the target appearance variations.
  \item  A novel trajectory rectification strategy is proposed that can be readily extended to other tracking algorithms to ensure more accurate and stable tracking results.
  \item  Our tracker achieves favorable results with $50.4\%$ in success plots and $71.4\%$ in precision plots, which ranks 1st in the CVPR2013 tracking benchmark~\cite{wu2013online}, showing the power of the simple Haar-like features.
\end{itemize}
\begin{figure}[t]
  \centering
  \includegraphics[width=1\linewidth]{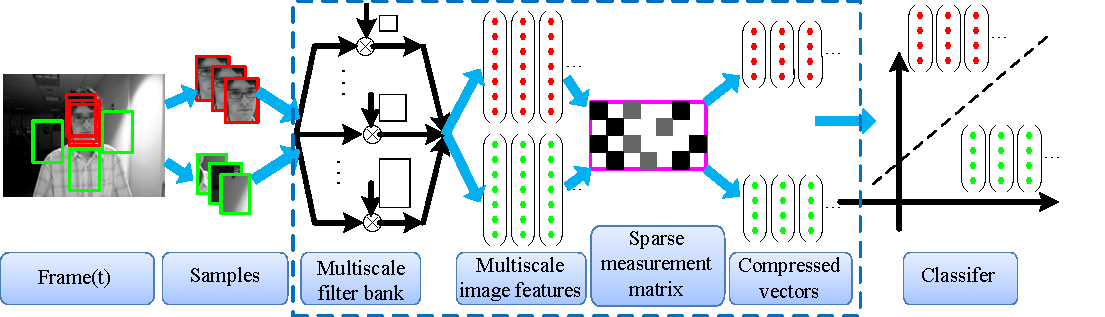}\\
  \caption{The main components of CT}
  \label{fig:procedure_ct}
\end{figure}
\section{Compressive Tracking}
The idea of CT~\cite{zhang2012real} is motivated by the compressive sensing theory~\cite{candes2005decoding,candes2006near} in which the random projections of a sufficiently high dimensional feature vector contain enough information to reconstruct the original high-dimensional one.
The main components are illustrated by Figure~\ref{fig:procedure_ct}. First, each sample is represented by a high-dimensional multiscale vector via convolving each patch  with some rectangle filters. Then, the vector is projected onto a low-dimensional space with a very sparse random projection matrix that satisfies the restricted isometry property (RIP) of the compressive sensing theory. The original high-dimensional feature vectors can well discriminate the target from its local background while the high efficiency is achieved by the very sparse random matrix, and thereby CT performs well on some challenging sequences in terms of both efficiency and accuracy.

CT employs a very sparse random matrix $\mathbf{R} \in {\mathbb{R}^{n \times {{(wh)}^2}}}$ to project the high-dimensional vector $\mathbf{x}$ onto a low-dimensional vector $\mathbf{v} \in {\mathbb{R}^n}$
\begin{equation}
\label{eq:randompj}
\mathbf{v} = \mathbf{R}\mathbf{x},
\end{equation}
where the entry of $\mathbf{R}$ is represented by
\begin{equation}
r_{ij}=\sqrt{\rho} \times \left\{\begin{array}{rl}
1 & \ \mbox{with probability } \frac{1}{2\rho}\\
0 & \ \mbox{with probability } 1-\frac{1}{\rho}\\
-1& \ \mbox{with probability } \frac{1}{\rho}.
\end{array} \right.
\label{eq:randommatrix}
\end{equation}
where $\rho = \frac{{{{(w \times h)}^2}}}{4}$ with $w$ and $h$ representing width and height of the object size, respectively.

\begin{figure}[tb]
  \centering
  \includegraphics[width=0.35\linewidth]{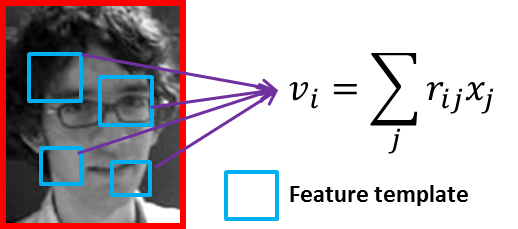}\\
  \caption{Each compressed feature is constructed by several feature templates}
  \label{fig:featuretemplate}
\end{figure}
In~(\ref{eq:randompj}), the $i$-th feature $v_i$ in the compressed vector $\mathbf{v}$ can be represented as
\begin{equation}
\label{eq:vi}
v_i=\sum_{j}r_{ij}x_{j}.
\end{equation}

Figure~\ref{fig:featuretemplate} illustrates that $v_i$ in (\ref{eq:vi}) is constructed by several feature templates, whose sizes and locations are set randomly and fixed during tracking. Although this random selection strategy is simple and efficient, it has some limitations that makes CT perform unfavorably when the target appearance varies much: First, the feature templates may select noninformative features when they fall into the textureless regions. Second, the fixed templates cannot adapt to the target appearance variations well. In the following section, we will propose an adaptive CT that can deal with these issues well.
\begin{figure}[t]
  \centering
  \includegraphics[width=1\linewidth]{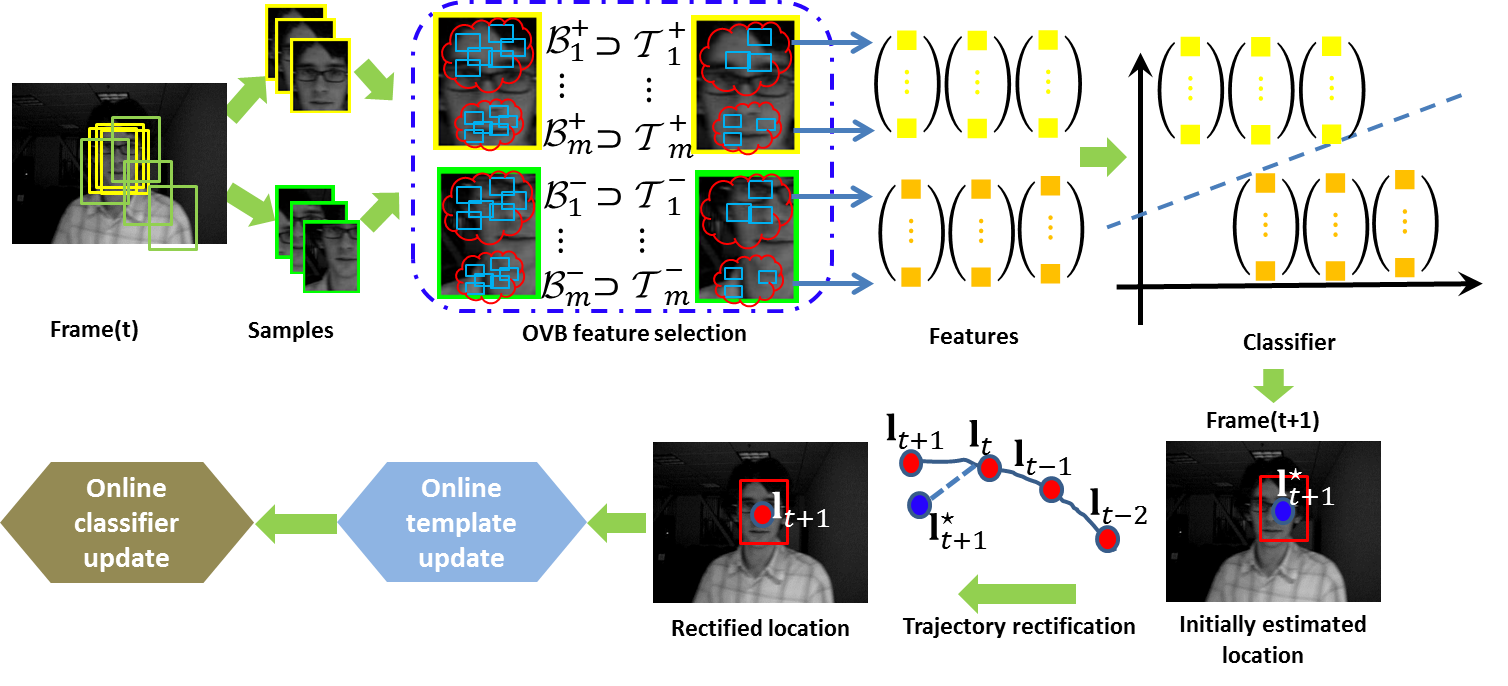}\\
  \caption{Flowchart of ACT. OVB is short for online vector boosting (refer to Algorithm~\ref{alg:OVB}).}
  \label{fig:act}
\end{figure}
\begin{algorithm}[t]
\caption{Adaptive Compressive Tracking (ACT)}
\label{alg:act}
\begin{algorithmic}[1]
\REQUIRE The $t$-th image frame
\STATE Sample a set of image patches in $D^{\gamma}=\{\textbf{p}|||\mathbf{l}_{t}(\textbf{p})-\mathbf{l}_{t-1}||<\gamma\}$
 where ${\mathbf{l}_{t - 1}}$ is the tracking location at the $(t-1)$-th frame, and extract features with the feature template $\mathcal{T}$
\STATE Apply the classifier $H(\cdot)$ in (\ref{eq5}) to each feature vector, and get the maximum confidence score $conf$
\IF{$conf < \Theta$}
\STATE Rectify the tracking location ${\mathbf{l}_t}$ via~(\ref{eq:trajectory})
\ELSE
\STATE Find the tracking location ${\mathbf{l}_t}$ via maximizing the classification score
\ENDIF
\STATE Sample two sets of image patches $D^\alpha=\{\textbf{p}|||\mathbf{l}_t(\textbf{p})-\mathbf{l}_t||<\zeta\}$ and $D^{\alpha,\beta}=\{\textbf{p}|\alpha<||\mathbf{l}_t(\textbf{p})-\mathbf{l}_t||<\beta\}$ with $\zeta<\alpha<\beta$
\STATE Update the feature template bags $\mathcal{B}$ and the classifier parameters
\ENSURE Tracking location ${\mathbf{l}_{t}}$
\end{algorithmic}
\end{algorithm}
\section{Adaptive Compressive Tracking}
\subsection{Algorithm overview}
Figure~\ref{fig:act} illustrates the basic flow of our tracking algorithm that is summarized in Algorithm~\ref{alg:act}, which mainly consists of three steps. First, we construct a set of positive and negative feature template bags $\{\mathcal{B}_i^+,\mathcal{B}_i^-\}_{i=1}^c$, in which each bag $\mathcal{B}_i=\{\textbf{z}_{ij}\}_{j=1}^n$ contains $n$ rectangle feature templates, of which each template $\textbf{z}_{ij}$ represents a vectorized image patch inside the blue rectangle, and then we select $k$ templates via an online vector boosting feature selection strategy, which constructs the feature template bags $\mathcal{T}_i\subset\mathcal{B}_i$. Second, to take into account the target appearance variations over time, we exploit an online template update strategy that preserves the stable feature templates while replacing the ones with remarkable changes by a linear combination of former and current templates. Finally, when the score of the maximum classifier confidence for the estimated tracking location is lower than a threshold $\Theta$, which indicates that the estimation is inaccurate, then we employ a trajectory rectification strategy that utilizes the former tracking motion information to predict the current tracking location.
\begin{figure}[t]
  \centering
  \includegraphics[width=.8\linewidth]{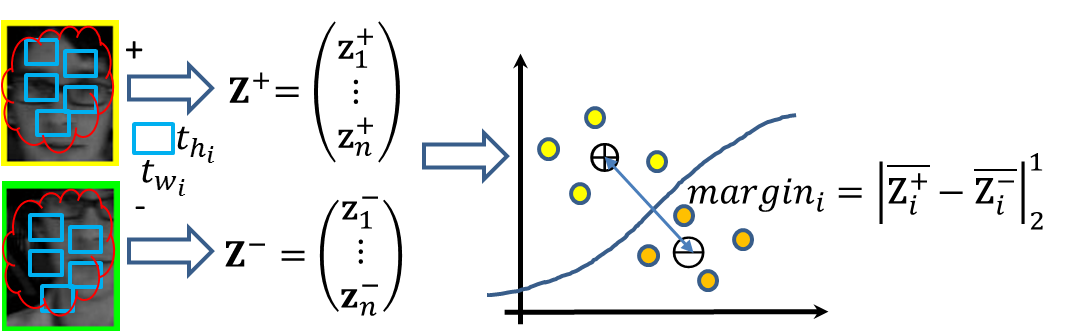}\\
  \caption{Illustration of the defined margin for the features in the $i$-th bag $\mathcal{B}_i$. $t_{w_i},t_{h_i}$ denote the width and height of the rectangle template in the $i$-th bag, respectively. $\mathbf{Z}_i^+,\mathbf{Z}_i^-$ denote the corresponding image representations of positive and negative samples, respectively, in which $\mathbf{z}_{i,1,\ldots,n}^+$ and $\mathbf{z}_{i,1,\ldots,n}^-$ denote the feature templates that are the normalized image patch vectors in the blue rectangles. $\overline{\mathbf{Z}_i^+}$ and $\overline{\mathbf{Z}_i^-}$ denote the average image representations of the positive and negative samples, respectively.}
  \label{fig:margin}
\end{figure}
\subsection{Online vector boosting feature selection}
As illustrated by Figure~\ref{fig:featuretemplate}, the templates in CT~\cite{zhang2012real} are constructed by several patches with random locations and sizes, of which the size of each patch ranges from $1\times1$ to $w\times h$ pixels, where $w$ and $h$ represent width and height of the object, respectively. However, a too small patch is vulnerable to the noisy small appearance variations while a too large one cannot distinguish the most likely target from its neighboring counterparts due to its large support. To handle this problem, we constrain the width and height of the feature template by $2 < t_{w_i} < round(w/2)$, $2 < t_{h_i} < round(h/2)$. Furthermore, to take into account multiscale appearance information, we set the patches in the same bag to the same size while the patches in different bags own varying sizes. Next, we will introduce our OVB feature selection method that can select the most discriminative feature templates from each bag to design a strong classifier.

As illustrated by Figure~\ref{fig:margin}, providing the positive and negative feature template bags $\mathcal{B}_i^+,\mathcal{B}_i^-,i=1,\ldots,c$, we define
a margin between them that is the sum of Euclidean distance between the average positive and negative feature vectors in each template
\begin{equation}
\label{eq:margin}
margin=\sum_{i=1}^c margin_i,
\end{equation}
where $margin_i$ is defined as
\begin{equation}
\begin{aligned}
\label{eq:margini}
margin_i&=|\overline{\mathbf{Z}_i^+}-\overline{\mathbf{Z}_i^-}|_2^1\\
        &=\sqrt{{\overline{\mathbf{Z}_i^+}}^\top\overline{\mathbf{Z}_i^+}-{\overline{\mathbf{Z}_i^+}}^\top\overline{\mathbf{Z}_i^-}-{\overline{\mathbf{Z}_i^-}}^\top\overline{\mathbf{Z}_i^+}+{\overline{\mathbf{Z}_i^-}}^\top\overline{\mathbf{Z}_i^-}}\\
        &=\sqrt{2n-2\sum_{j=1}^n{\overline{\mathbf{z}_{ij}^+}}^\top\overline{\mathbf{z}_{ij}^-}}
\end{aligned}
\end{equation}
where $\overline{\mathbf{z}_{ij}^+}$ and $\overline{\mathbf{z}_{ij}^-}$ denote the $j$-th normalized feature templates in the $i$-th bag of the positive and negative samples, respectively.

It is easy to verify that $\overline{\mathbf{z}_{ij}^+}\top\overline{\mathbf{z}_{ij}^-}\leq\overline{\mathbf{z}_{ij}^+}\top\sum_{j=1}^n\overline{\mathbf{z}_{ij}^-}$, so we have
\begin{equation}
margin_i=\sqrt{2n-2\sum_{j=1}^n{\overline{\mathbf{z}_{ij}^+}}^\top\overline{\mathbf{z}_{ij}^-}}\geq\sqrt{2n-2\sum_{j=1}^n{\overline{\mathbf{z}_{ij}^+}}^\top\sum_{j=1}^n\overline{\mathbf{z}_{ij}^-}}
=\mathcal{J}(\overline{\mathbf{z}_{i1}}+,\ldots,+\overline{\mathbf{z}_{in}}),
\end{equation}
where $\mathcal{J}$ is the lower bound of the margin function $margin_i$. For each sample $\mathbf{p}$, its image representation in the $i$-th bag is $\mathcal{B}_i(\mathbf{p})=\{\mathbf{z}_{ij}(\mathbf{p})\}_{j=1}^n$, and we utilize the template center bag to robustly represent each class as $\overline{\mathcal{B}_i}=\{\overline{\mathbf{z}_{ij}}\}_{j=1}^n$ (See Figure~\ref{fig:margin}). Our objective is to select a subsect of feature templates $\{\overline{\mathbf{z}_{ij}}\}_{j=1}^k$ from bag $\overline{\mathcal{B}_i}$ that maximizes the margin function $margin_i$, which can be readily achieved by maximizing its lower bound $\mathcal{J}$
\begin{equation}
\label{eq:optimizeJ}
\{\overline{\mathbf{z}_{i1}},\ldots,\overline{\mathbf{z}_{ik}}\}={\arg\max}_{\{\overline{\mathbf{z}_{i1}},\ldots,\overline{\mathbf{z}_{ik}}\}\subset\overline{\mathcal{B}_i}}\mathcal{J}(\overline{\mathbf{z}_{i1}}+,\ldots,+\overline{\mathbf{z}_{ik}}).
\end{equation}
The vector boosting algorithm in~\cite{huang2005vector} relies on the special case of the exponential loss function of AdaBoost, and thus cannot be readily adapted to solve the above problem. Now, we present the proposed novel online vector boosting algorithm that can readily address the above problem. Our method is motivated by the algorithm in~\cite{friedman2000additive} that takes the statistical view of boosting, which tries to optimize a specific objective function $\mathcal{L}$ by sequentially optimizing the following criterion
\begin{equation}
\label{eq:L}
(h_j,\alpha_j)={\arg\max}_{h_j\in \mathcal{H},\alpha}\mathcal{L}(H_{j-1}+\alpha h_j),
\end{equation}
where $H_{j-1}(\textbf{p}):\Omega\rightarrow \mathbb{R}$ is a strong classifier that is the sum of the first $j-1$ weak classifiers $h_i,i=1,\ldots,j-1$ and $\mathcal{H}$ is the set of all possible weak classifiers.

The proposed algorithm is an extension of the optimization problem in~(\ref{eq:L}) in which both the outputs of its weak classifiers and final output are vectors rather than scalars.
At all time we maintain $n$ candidate weak classifiers in which the $j$-th weak classifier is defined as
\begin{equation}
\label{eq:hij}
\mathbf{h}_{ij}(\mathbf{p})=\overline{\mathbf{z}_{ij}}(\mathbf{p}).
\end{equation}
To update the classifier, we first update a subset of weak classifiers in parallel via an online feature template update strategy (refer to Section~\ref{sec:templateupdate}), and then we choose $k$ weak classifiers $\mathbf{h}_{ij}$ from the candidate pool sequentially by maximizing the lower bound $\mathcal{J}$ in~(\ref{eq:optimizeJ})
\begin{equation}
\mathbf{h}_{ij} = {\arg\max}_{\mathbf{h}_{ij}\in\{\overline{\mathbf{z}_{i1}},\ldots,\overline{\mathbf{z}_{in}}\}}\mathcal{J}(\mathbf{H}_{i{(j-1)}}+\mathbf{h}_{ij}),
\end{equation}
where $\mathbf{H}_{i{(j-1)}}=\sum_{l=1}^{j-1}\mathbf{h}_{il}$.
Algorithm~\ref{alg:OVB} summarizes the main steps of the proposed online vector boosting method.

\begin{algorithm}
\caption{Online Vector Boosting (OVB)}
\label{alg:OVB}
\begin{algorithmic}
\label{alg2}
\STATE \textbf{Input:} Feature templates $\{\mathbf{z}_{ij}^+,\mathbf{z}_{ij}^-,j=1,\ldots,n\}$.
\begin{enumerate}\setlength{\itemsep}{-\itemsep}
\item Update $n$ weak classifiers $\mathbf{h}_{ij},j=1,\ldots,n$ according to the strategy introduced by Section~\ref{sec:templateupdate}
\item Initialize $\mathbf{H}_{ij}=\mathbf{0}$ for all $i,j$
\item \textbf {for} $j=1$ to $k$ \textbf{do}
\item  \quad \textbf{for} $m=1$ to $n$ \textbf{do}
\item \quad\quad $\mathcal{J}^m=\mathcal{J}(\mathbf{H}_{ij}+\mathbf{h}_{im})$
\item \quad \textbf{end for}
\item \quad$m_j^\star= {\arg\max}_m\mathcal{J}^m$
\item \quad $\mathbf{h}_{ij}\leftarrow \mathbf{h}_{im_j^\star}$
\item \quad $\mathbf{H}_{ij}=\mathbf{H}_{ij}+\mathbf{h}_{ij}$
\item \textbf{end for}
\end{enumerate}
\STATE\textbf{Output:} $k$ selected feature templates $\{\mathbf{z}_{im_j^\star}^+,\mathbf{z}_{im_j^\star}^-,j=1,\ldots,k\}$
\end{algorithmic}
\end{algorithm}
At last, we concatenate all the selected feature templates in all bags to yield a high-dimensional multiscale image representation $\mathbf{x}=(\mathbf{h}_{11}^\top,\ldots,$ $\mathbf{h}_{1k}^\top,\ldots,\mathbf{h}_{c1}^\top,\ldots,\mathbf{h}_{ck}^\top)^\top\in\mathbb{R}^{k\sum_{i=1}^ct_{w_i}t_{h_i}\times 1}$. We then utilize an orthogonal matrix $\mathbf{S}\in \mathbb{R}^{c\times k\sum_{i=1}^ct_{w_i}t_{h_i}}$ to project $\mathbf{x}$ onto a $c$-dimensional feature space
\begin{equation}
\mathbf{v}=\mathbf{Sx},
\end{equation}
where the entry of $\mathbf{S}$ is denoted as
\begin{equation}
s_{ij}=\frac{1}{\sqrt{kt_{w_i}t_{h_i}}} \times \left\{\begin{array}{rl}
0 & \ j<(i-1)kt_{w_i}t_{h_i},j>ikt_{w_i}t_{h_i},\\
\pm 1 & \ \mbox{with equal probability,}\\
\end{array} \right.
\label{eq:projectS}
\end{equation}
and the $i$-th entry of $\mathbf{v}$ is represented as
\begin{equation}
v_i = \sum_{j=(i-1)\times c+1}^{(i-1)\times c+k+1}s_{ij}sum(\mathbf{h}_{ij}),
\end{equation}
where $sum(\mathbf{h}_{ij})$ can be efficiently computed by the integral images.
\subsection{Online feature template update}
\label{sec:templateupdate}
CT~\cite{zhang2012real} suffers drift when the target appearance changes much due to its fixed feature templates. In our algorithm, we proposes a conservative update scheme that only updates the templates with significant variations. Let $\Delta_{ij}=|\mathbf{h}_{ij}(\mathbf{p}_t)-\mathbf{h}_{ij}(\mathbf{p}_{t-1})|_2^1$ denote the corresponding template variations between two consecutive frames. If $\Delta_{ij}<\vartheta$, we keep the template $\mathbf{h}_{ij}$, otherwise, we update the template
\begin{equation}
\mathbf{h}_{ij}(\mathbf{p}_t)=\eta \mathbf{h}_{ij}(\mathbf{p}_t)+(1-\eta)\mathbf{h}_{ij}(\mathbf{p}_{t-1}).
\end{equation}
where $\eta>0$ represents the update ratio.

\subsection{Online trajectory rectification}
Similar to CT~\cite{zhang2012real}, the tracking task is treated as a binary classification problem that the Naive Bayes classifier is adopted
\begin{equation}\label{eq5}
H(\mathbf{v}) = \log\left(\frac{{\prod\nolimits_{i = 1}^c {p({v_i}|y = +)} p(y = +)}}{{\prod\nolimits_{i = 1}^c {p({v_i}|y = -)} p(y = -)}}\right) = \sum\limits_{i = 1}^c {\log \left(\frac{{p({v_i}|y = +)}}{{p({v_i}|y = -)}}\right),}
\end{equation}
and the conditional distributions are assumed to be Gaussian distributed as
\begin{equation}\label{eq6}
p({v_i}|y = +) \sim \mathcal{N}(\mu_i^+,\sigma_i^+),p({v_i}|y = -) \sim \mathcal{N}(\mu_i^-,\sigma_i^-),
\end{equation}
where $\mu_i^+$ and $\sigma_i^+$ are the mean and standard deviation of the $i$-th positive feature, respectively and similar to $\mu_i^-$ and $\sigma_i^-$.
The parameters $\mu_i^+$ and $\sigma_i^+$ are incrementally update by
\begin{equation}\label{eq7}
\begin{array}{l}
\mu_i^+ \leftarrow \lambda \mu_i^+ + (1 - \lambda )\mu^+,\\
\sigma_i^+ \leftarrow \sqrt {\lambda {{(\sigma_i^+)}^2} + (1 - \lambda ){{(\sigma^+)}^2} + \lambda (1 - \lambda ){{(\mu_i^+ - \mu^+)}^2}},
\end{array}
\end{equation}
where $0 < \lambda  < 1$ is a learning parameter, $\sigma^+=\sqrt{\frac{1}{n^+}\sum_{k=0|y=+}^{n^+-1}
(v_{i}(k)-\mu^+)^2}$ and $\mu^+=\frac{1}{n^+}\sum_{k=0|y=+}^{n^+-1}v_{i}(k)$, $n^+$ is the number of positive samples. Parameters  $\mu_{i}^{-}$ and $\sigma_{i}^{-}$ are updated with similar
rules.

When $conf=\max_\mathbf{v} H(\mathbf{v})<\Theta$, which means the maximum classifier response is determined by the negative samples, the templates stop update. Then, we utilize the motion status in the former consecutive frames to predict the object location
\begin{equation}
\label{eq:trajectory}
\mathbf{l}_t = \mathbf{l}_{t - \Delta t}+\overrightarrow{\mathbf{v}_t}\Delta t,
\end{equation}
where $\overrightarrow{\mathbf{v}_t}=\frac{\mathbf{l}_{t-1}-\mathbf{l}_{t-4}}{3}$ is the average motion velocity estimated from the former three frames, and $\Delta t =1$ is the time step.
\section{Experiments}
\subsection{Setup}
{\bf Dataset:} We evaluate the proposed algorithm on the CVPR2013 tracking benchmark~\cite{wu2013online} that includes results of $29$ tracking algorithms on 50 fully annotated sequences ($\sim 26000$ frames). For better evaluation and analysis of the strength and weakness of the tracking algorithms, the sequences are categorized according to $11$ attributes, including illumination variation (IV), scale variation (SV), occlusion (OCC), deformation (DEF), motion blur (MB), fast motion (FM), in-plane rotation (IPR), out-of-plane rotation (OPR), out-of-view (OV), background clutters (BC), and low resolution (LR).

{\bf Parameter setting:} The number of bags is set to $c=150$. The number of templates in each bag is set to $n=30$, in which $k=5$ templates are selected. Threshold of the classifier score $\Theta= 0$. Threshold of appearance update is set to $\vartheta=100$. The radius of searching window $\gamma=25$. The radius of sampling positive samples $\alpha=2$, where $n^+=40$ positive samples are extracted. The inner radius of sampling negative samples $\zeta=4$ while its corresponding outer radius $\beta=15$, where $n^-=40$ negative samples are selected. The update ratio of feature template $\eta=0.05$, and the learning parameter of the classifier is set to $\lambda=0.85$.

{\bf Evaluation metric:}
We employ the precision plot and  success plot defined in~~\cite{wu2013online} to evaluate the robustness of the tracking algorithms. The precision plot shows the percentage of frames whose estimated average center location errors are within the given threshold distance to the ground truth, in which the average center location error is defined as the average Euclidean distance between the center location of the tracked target and the manually labeled ground truth. The score at the threshold $20$ pixels is defined as the precision score. Success plot shows the percentage of successful frames at the threshold ranging from $0$ to $1$. The successful frame is defined as the overlap score more than a fixed value, where the overlap ratio is defined as $S = \frac{{\|{r_t}\bigcap {{r_a}}\| }}{{\|{r_t}\bigcup {{r_a}}\| }}$ with the tracking output bounding box ${r_t}$ and the ground truth bounding box ${r_a}$. For fair evaluation, the area under curve (AUC) is preferred to measure the success ratio. The one-pass evaluation (OPE) based on the average precision and the success rate given the ground truth of the first frame is used to evaluate the robustness of our algorithm.

\subsection{Quantitative comparisons}
\begin{figure}[!htb]
  \centering
  \includegraphics[width=.45\linewidth]{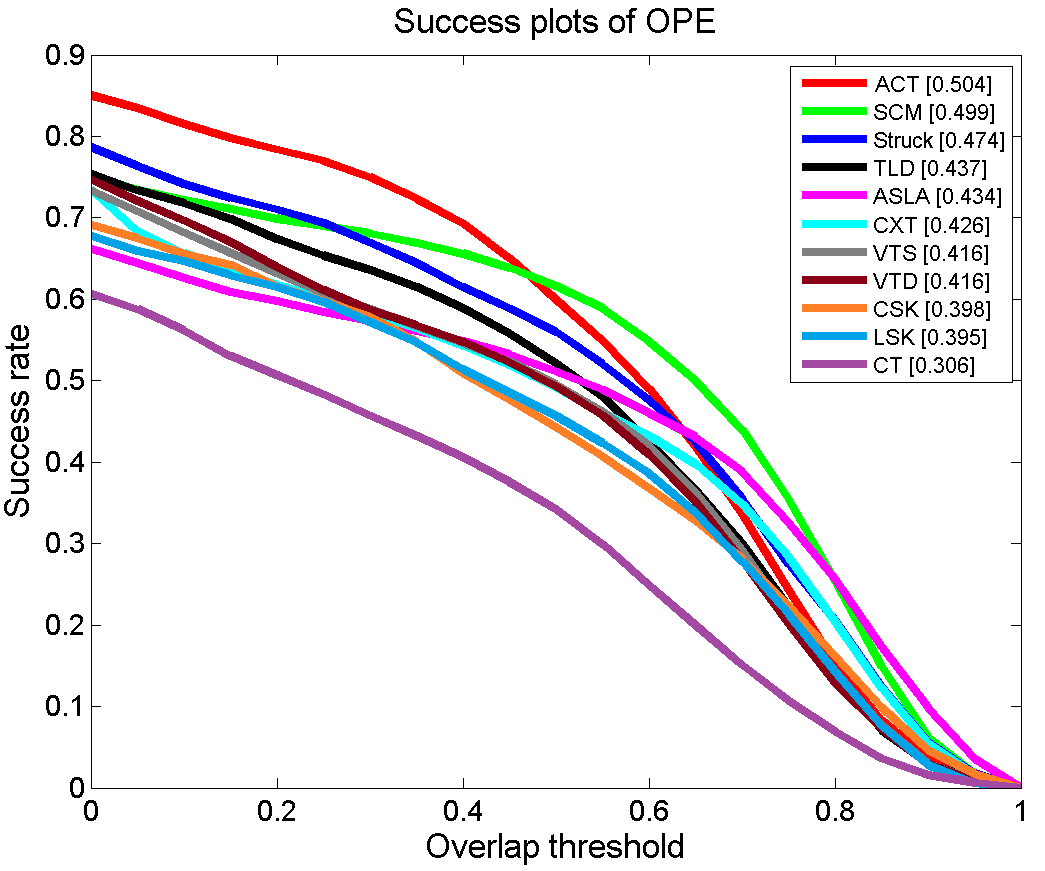}
  \includegraphics[width=.45\linewidth]{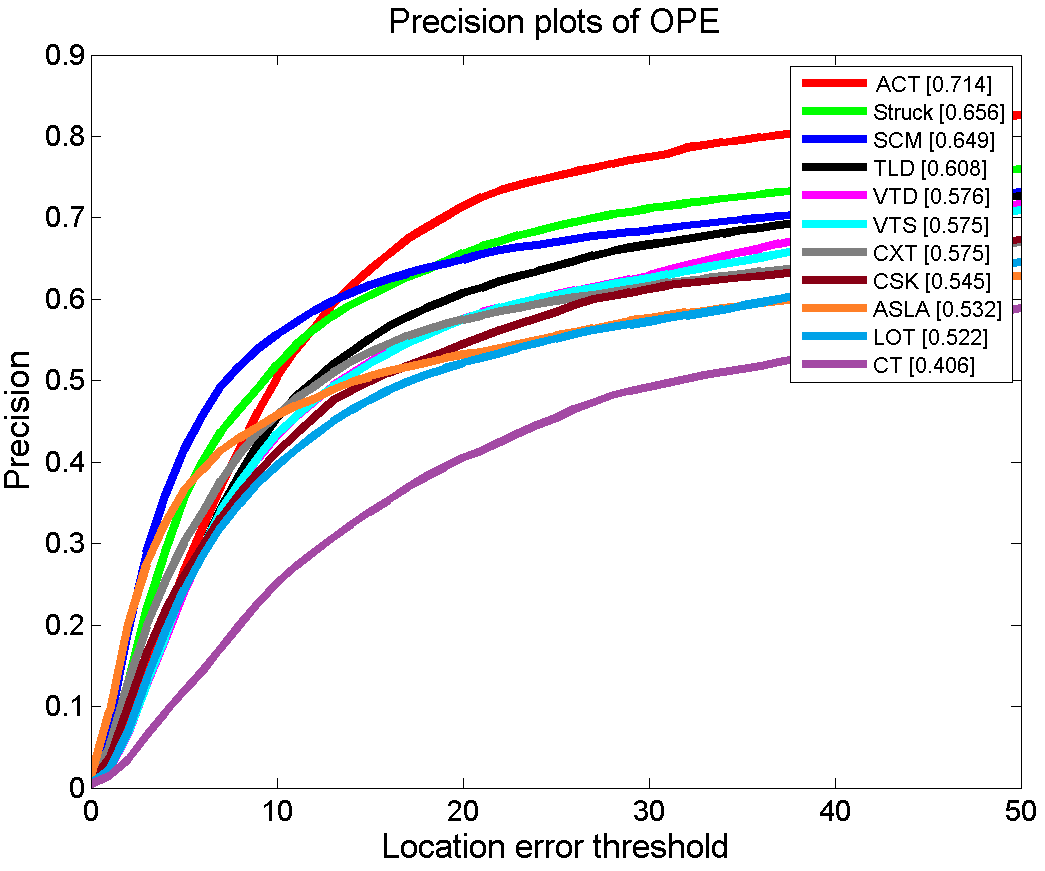}\\
  \caption{The success plots and precision plots of OPE for the top $10$ trackers and CT. The performance score for each tracker is shown in the legend. The performance score of precession plot is at the error threshold of $20$ pixels while the performance score of success plot is the AUC value. Best viewed on color display.}\label{fig5}
\end{figure}

{\bf Overall performance}: Figure~\ref{fig5} illustrates overall performance of the top $10$ evaluated tracking algorithms (i.e., SCM~\cite{jia2012visual}, Struck~\cite{hare2011struck}, TLD~\cite{kalal2010pn}, ASLA~\cite{zhong2012robust}, CXT~\cite{dinh2011context}, VTS~\cite{kwon2011tracking}, VTD~\cite{kwon2010visual}, CSK~\cite{henriques2012exploiting}, LSK~\cite{liu2011robust}, and LOT~\cite{oron2012locally}) and the CT algorithm~\cite{zhang2014fast} in terms of precision plot and success plot. The proposed ACT ranks 1st in terms of both precision plot and success plot: the precision score of ACT is $0.714$, which outperforms Struct~\cite{hare2011struck} ($0.656$); meanwhile, in the success plot, the proposed ACT achieves the score of AUC $0.504$, which outperforms SCM~\cite{zhong2012robust} by $0.5\% $. Note that the proposed ACT exploits only simple Haar-like features to represent the object and background, in which the simple naive Bayesian classifier is adopted with low computational complexity, yet it outperforms Struct and SCM that resort to complicate learning techniques in terms of both accuracy and efficiency.

{\bf Attribute-based performance}: To facilitate analyzing strength and weakness of the proposed algorithm, we further evaluate ACT on videos with $11$ attributes. Since the AUC score of the success plot is more accurate than that at the representative threshold (e.g., $20$ pixels) of the precision plot, in the following we mainly analyze ACT based on the success plot.

\begin{figure}[tb]
  \centering
  \includegraphics[width=.3\linewidth]{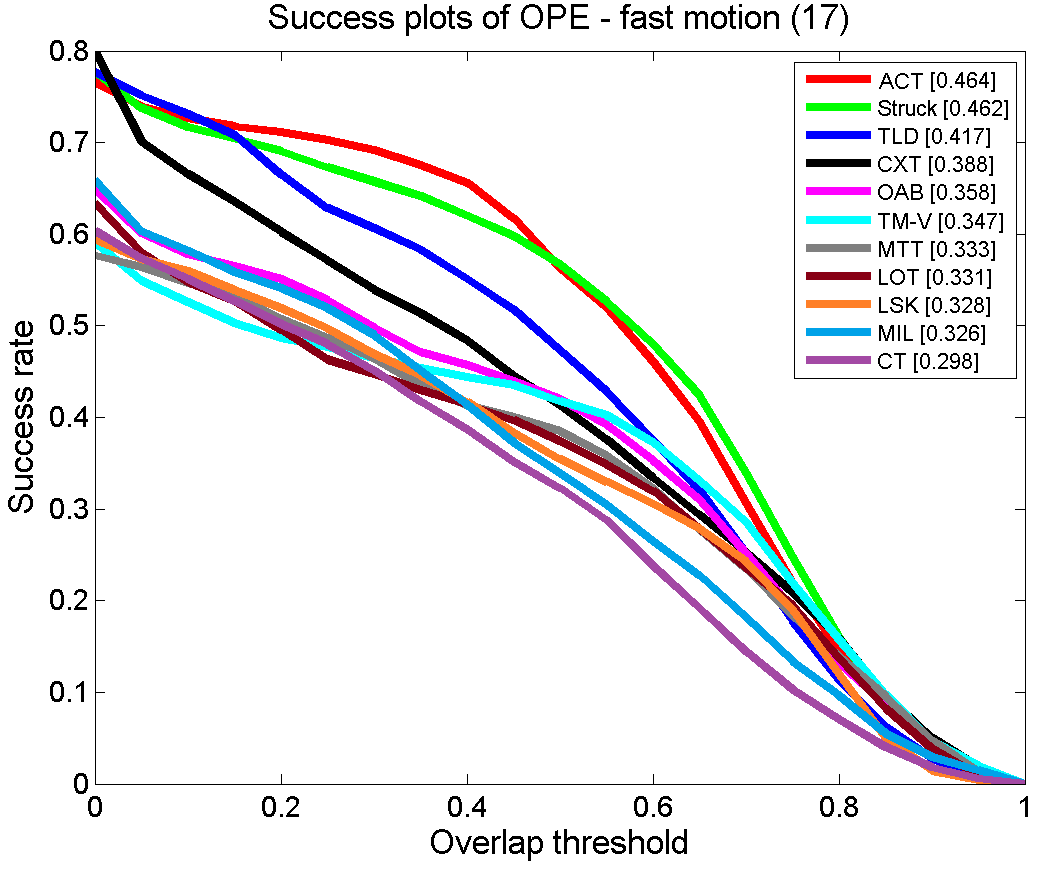}
  \includegraphics[width=.3\linewidth]{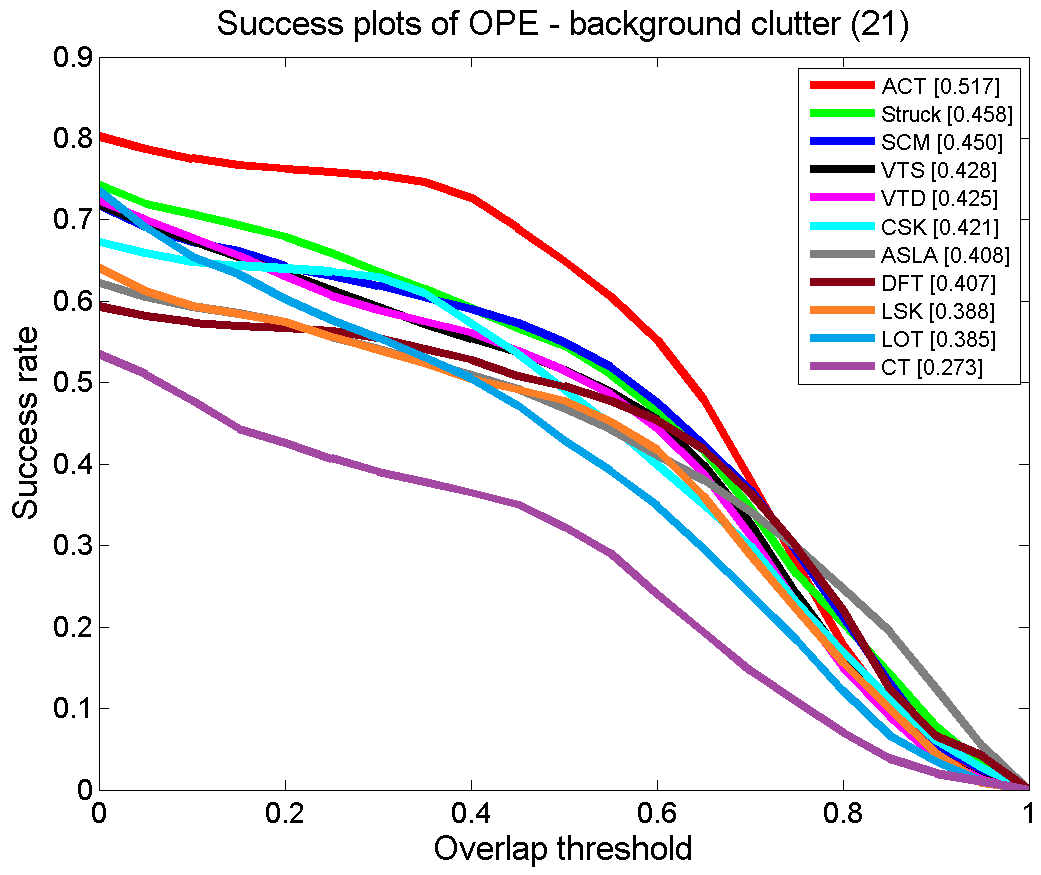}
  \includegraphics[width=.3\linewidth]{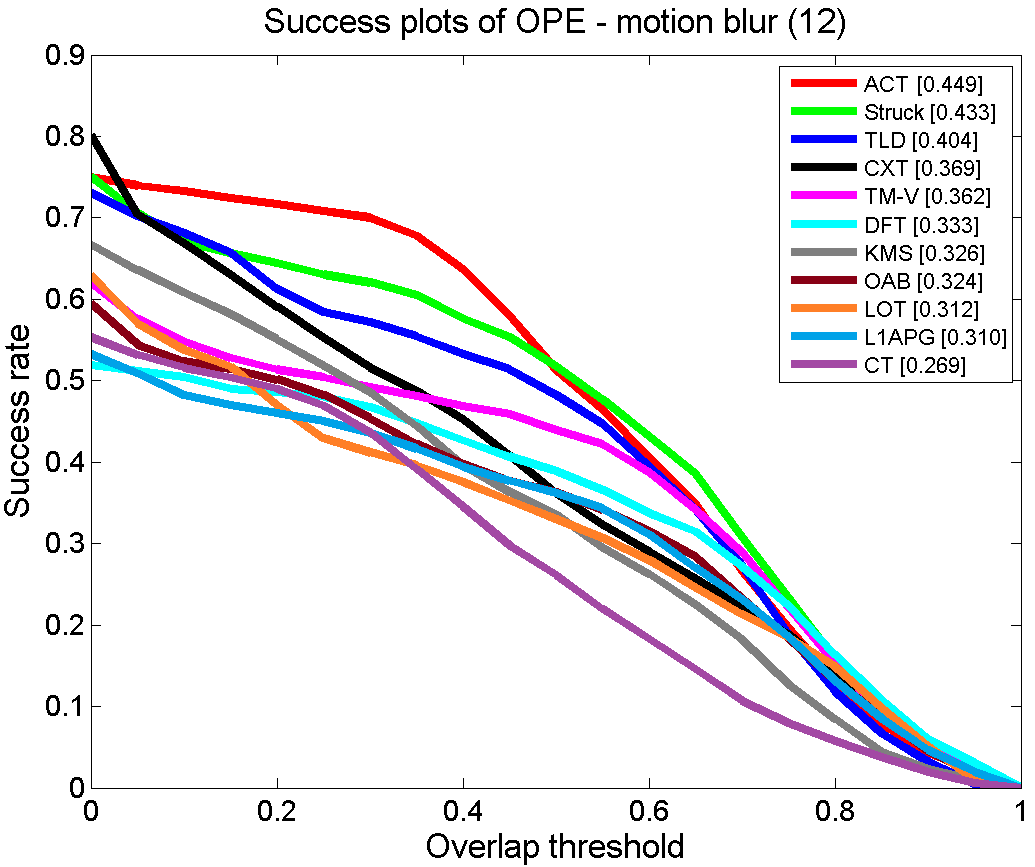}
  \includegraphics[width=.3\linewidth]{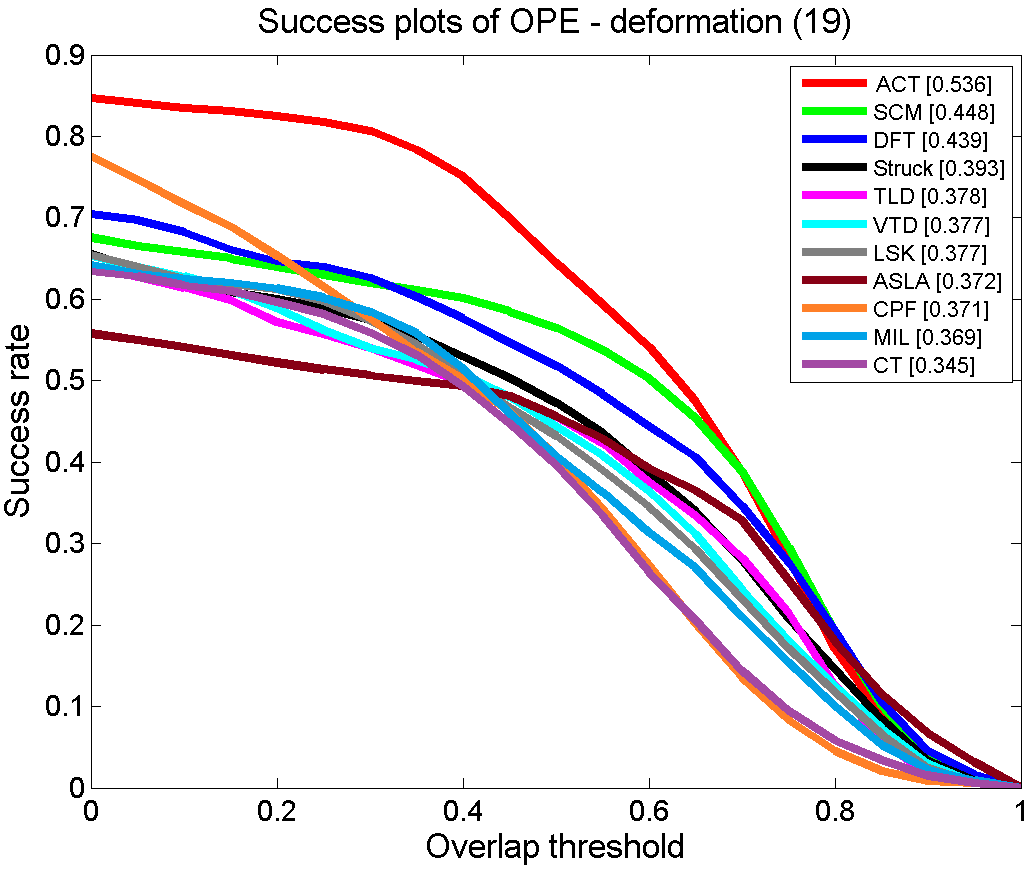}
  \includegraphics[width=.3\linewidth]{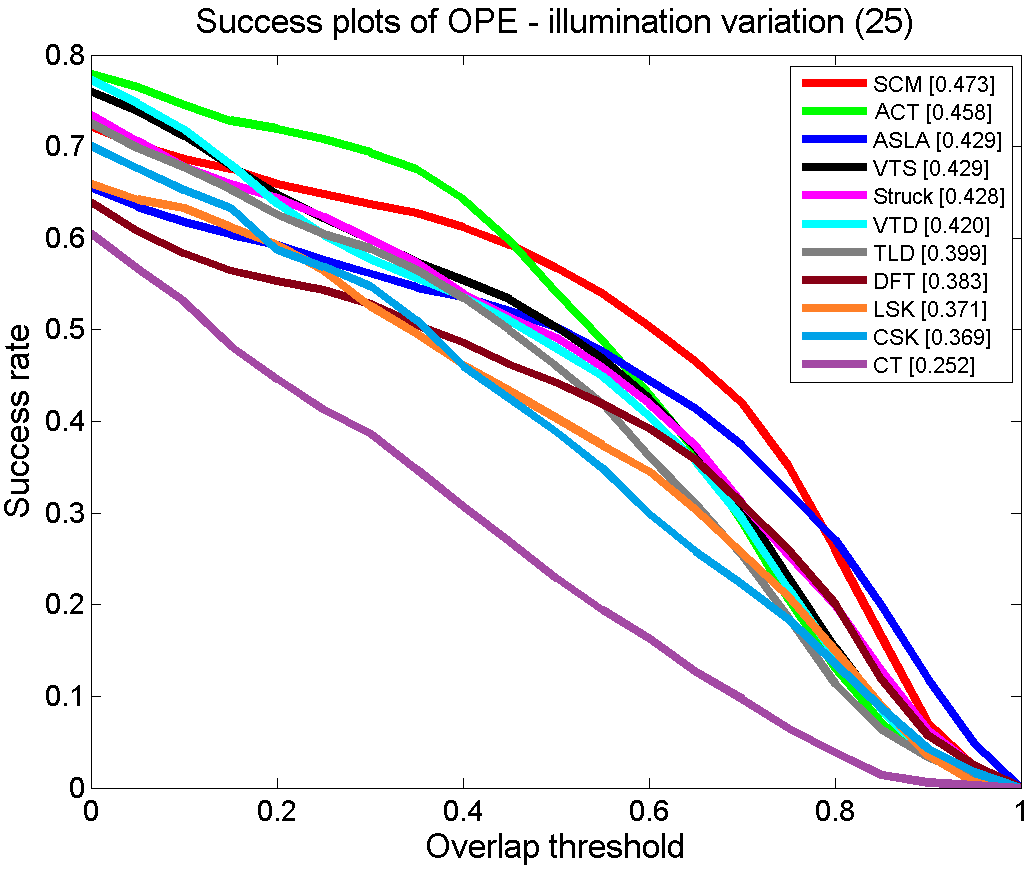}
  \includegraphics[width=.3\linewidth]{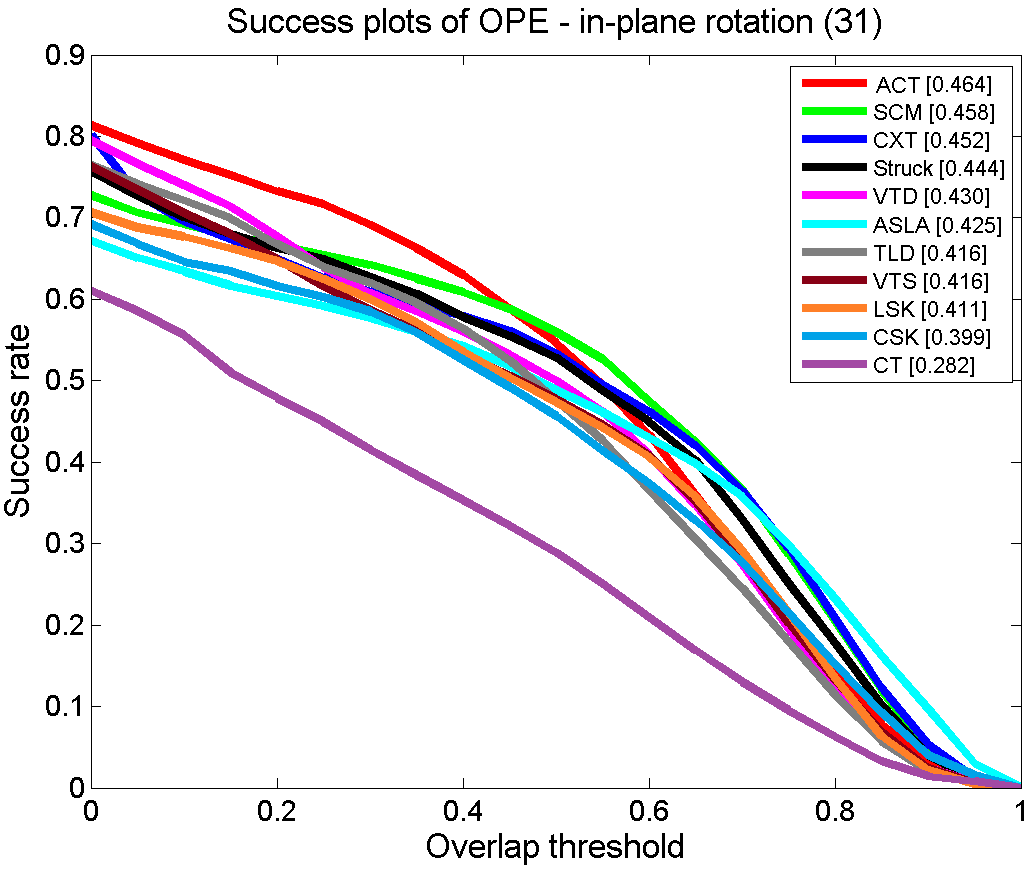}
  \includegraphics[width=.35\linewidth]{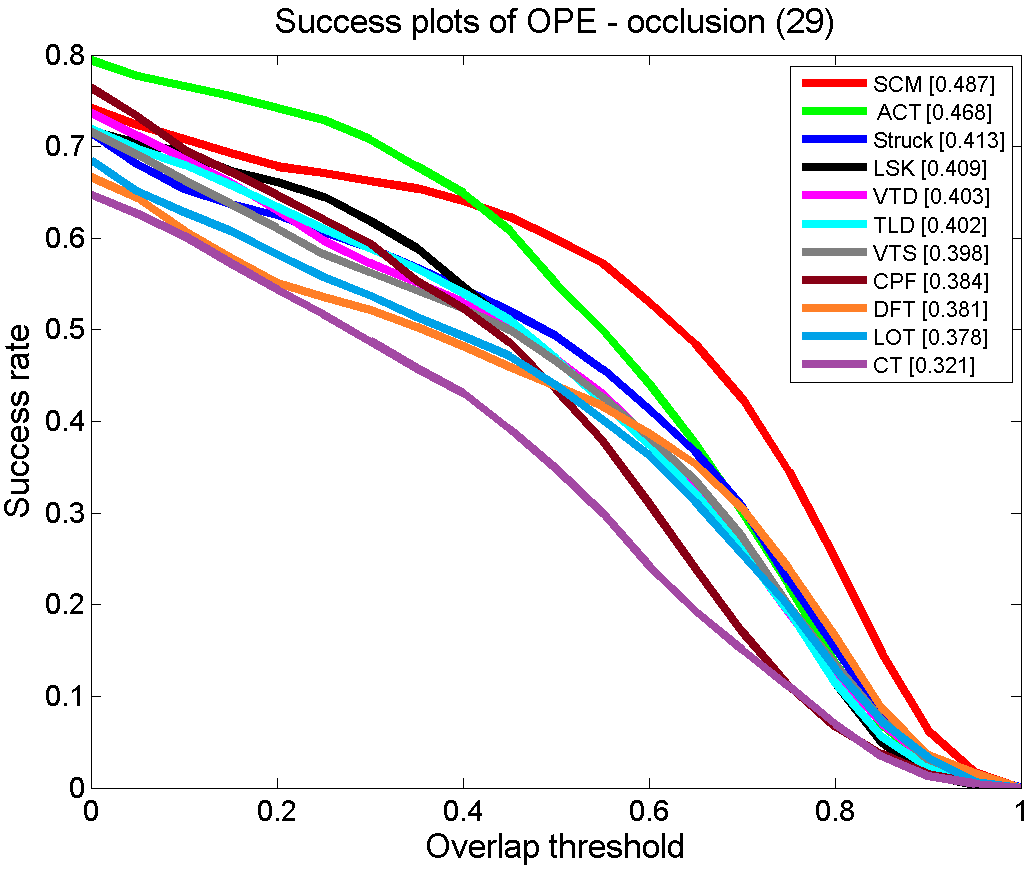}
  \includegraphics[width=.35\linewidth]{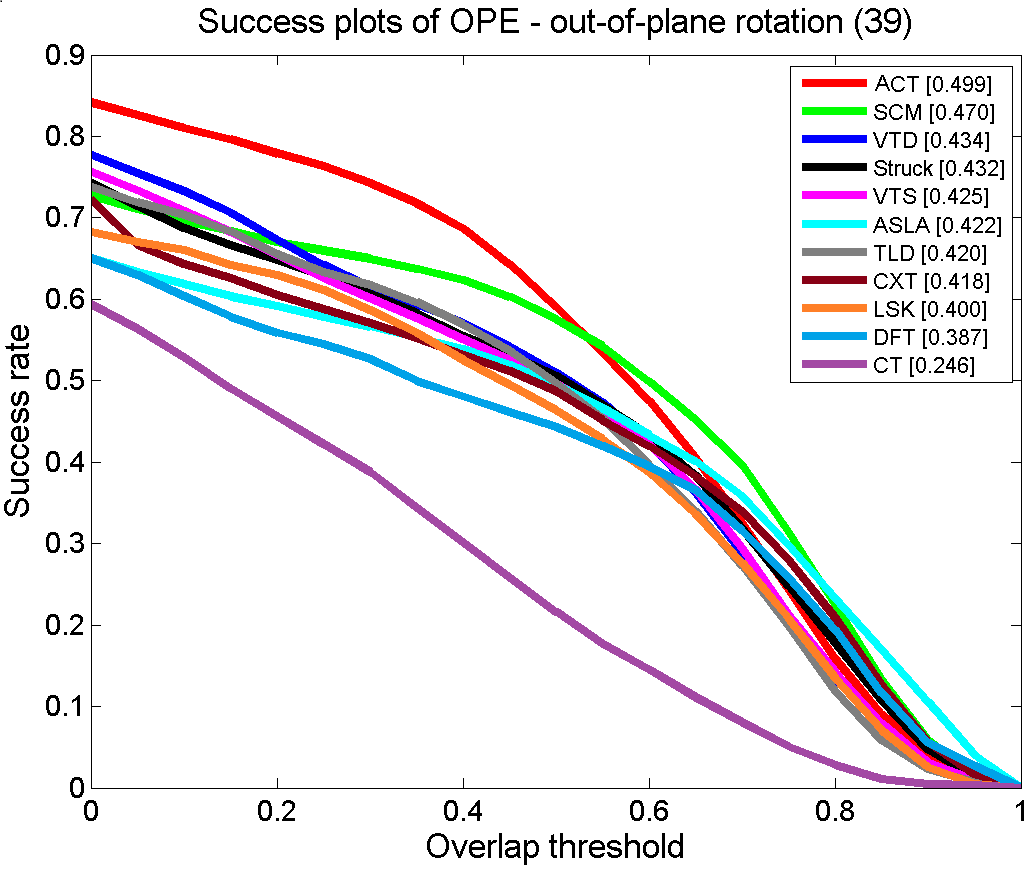}\\
  \caption{The success plots of videos with different attributes that ACT can achieve favorable results (within the top
$2$). Best viewed on color display.}\label{fig6}
\end{figure}

Figure~\ref{fig6} shows that the success plots of videos with attributes that our method achieves favorable results, in which ACT ranks within top $2$ on $8$ out of $11$ attributes. For the videos with attributes such as FM, MB, IV, DEF, BC, IPR, and OPR, ACT ranks 1st among all evaluated algorithms. In sequences with FM and MB, Struct ranks 2rd, showing that the tracker with wide range search window and dense sampling can perform well on these attributes, and so does ACT that sets search window size based on target size which prevents wrongly updating classifier from distracters. In sequences with IV and OCC, both SCM and ACT perform favorably well because they employ local features, in which ACT exploits the Haar-like features from the target via templates with varying sizes while SCM learns the local patch features from the target and background with sparse representation. Furthermore, both of them utilize the target template from the first frame, which is robust to drift problem. Similarly, on the OPR and IPR subsets, besides our tracker, the SCM and ASLSA trackers are also able to obtain satisfactory results, which can be attributed to the effective spare representations of local patches.

\begin{figure}[tbh]
  \centering
  \includegraphics[width=.3\linewidth]{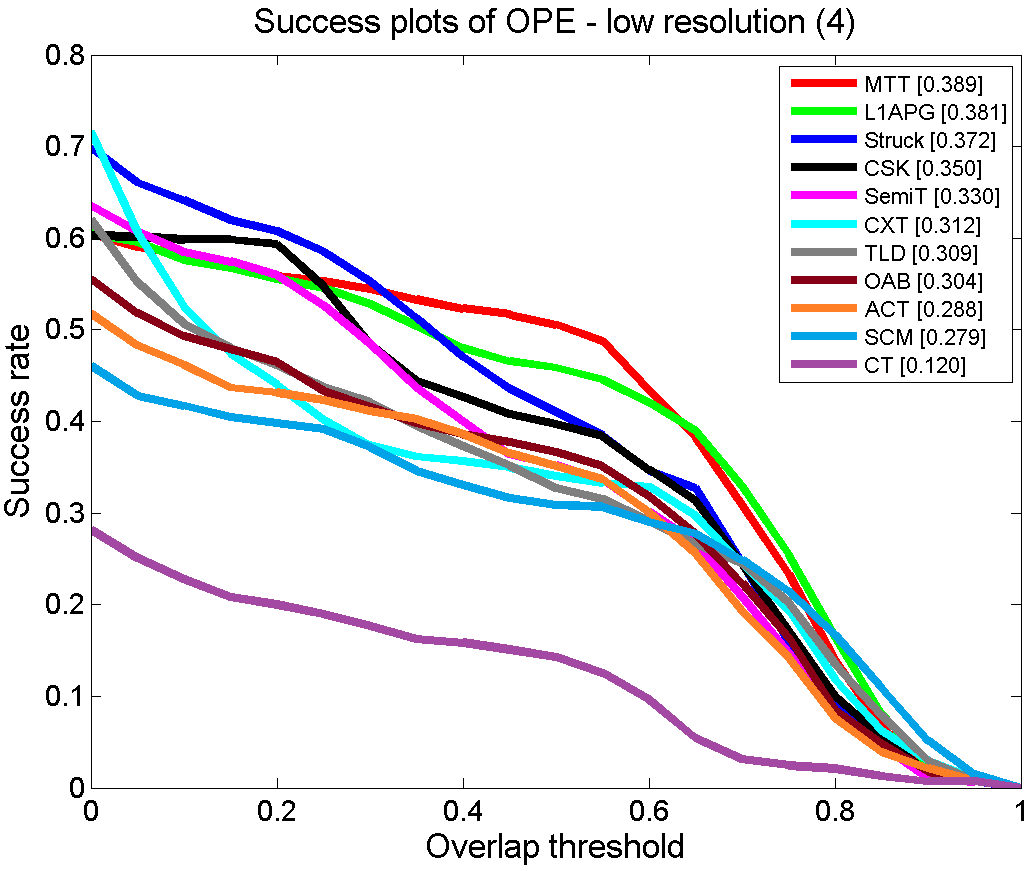}
   \includegraphics[width=.3\linewidth]{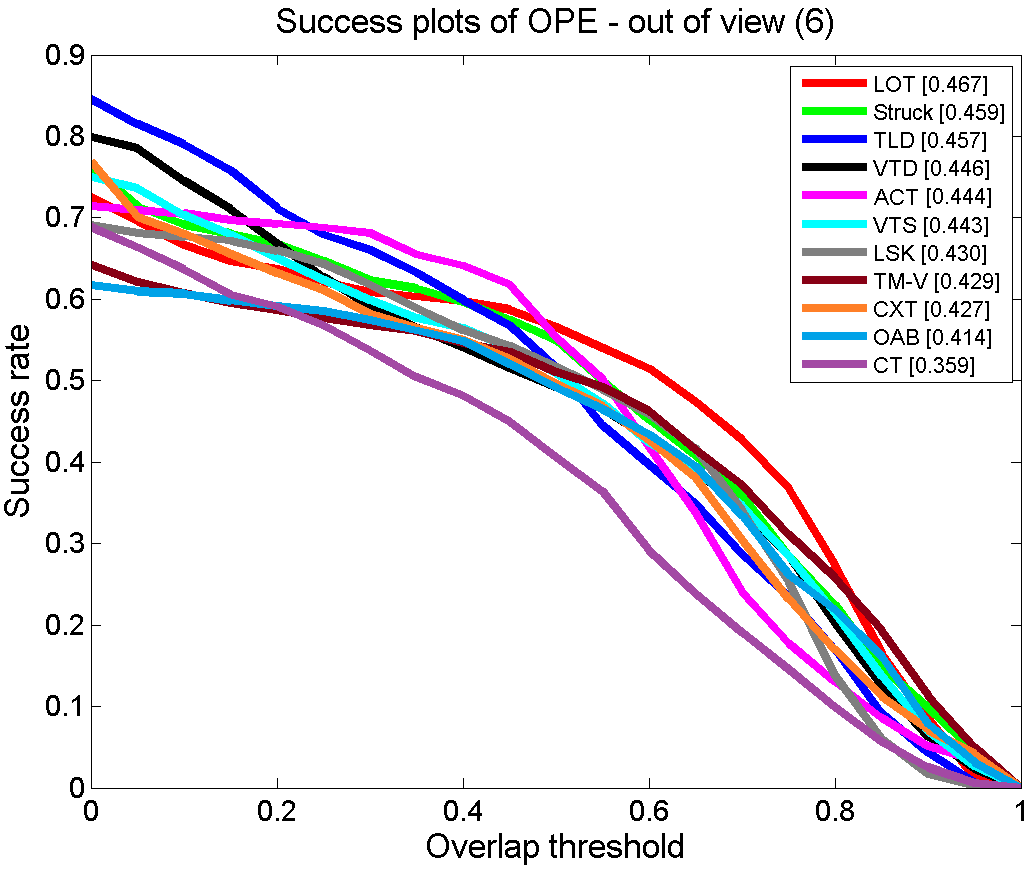}
    \includegraphics[width=.3\linewidth]{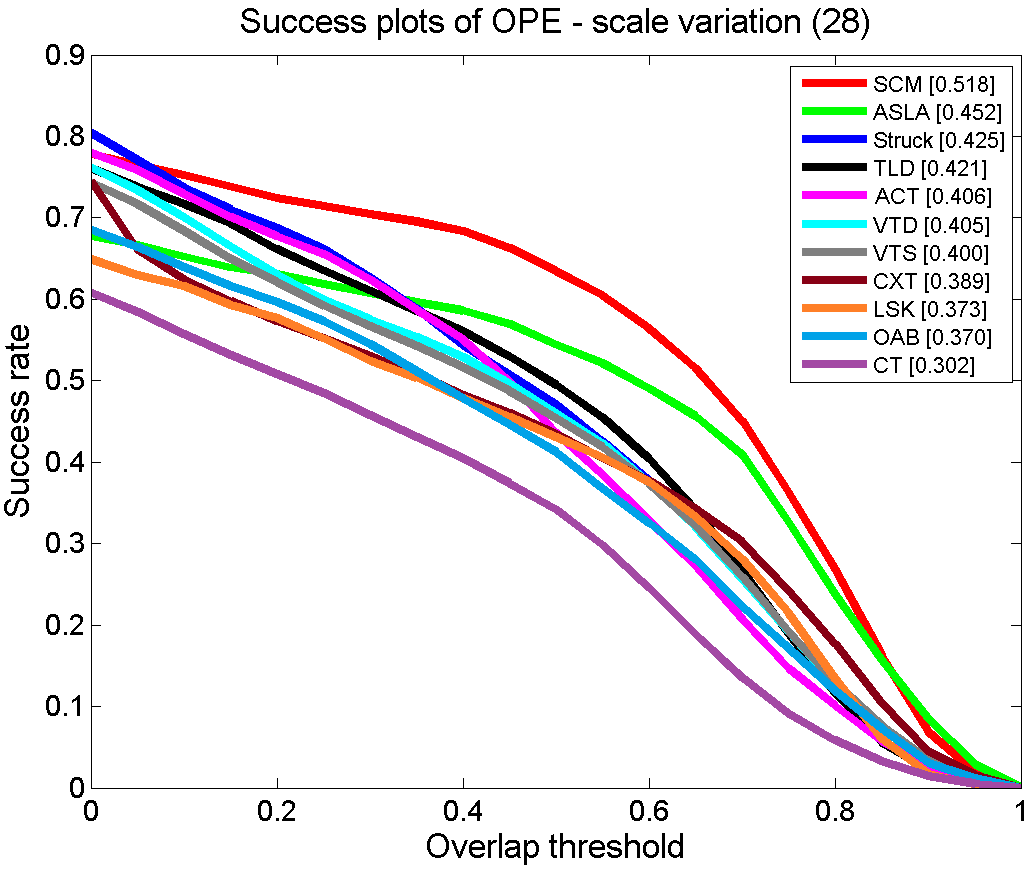}\\
  \caption{The success plots of videos with different attributes that ACT cannot perform well (outside the top 2). Best viewed on color display.}\label{fig7}
\end{figure}

Figure~\ref{fig7} shows that ACT cannot perform well with three attributes, such as LR, OV, and SV. The low resolution makes ACT less effective to extract useful information from the target object. This can be improved by considering the context information surrounding the target as~\cite{zhang2014fastv}. Furthermore, the target appearance change drastically when OV occurs, and thus ACT cannot deal with these drastic variation favorably. However, the tracking failure in this case can be well reduced by memorizing information from some former frames and enlarging the search range. Finally, ACT only considers single scale tracking for simplicity, which can be readily extended to the multi-scale tracking by constructing scale pyramids, thereby improving performance on the sequences with SV attribute.

\subsection{Qualitative comparisons}
\begin{figure}
  \centering
  \includegraphics[width=1\linewidth]{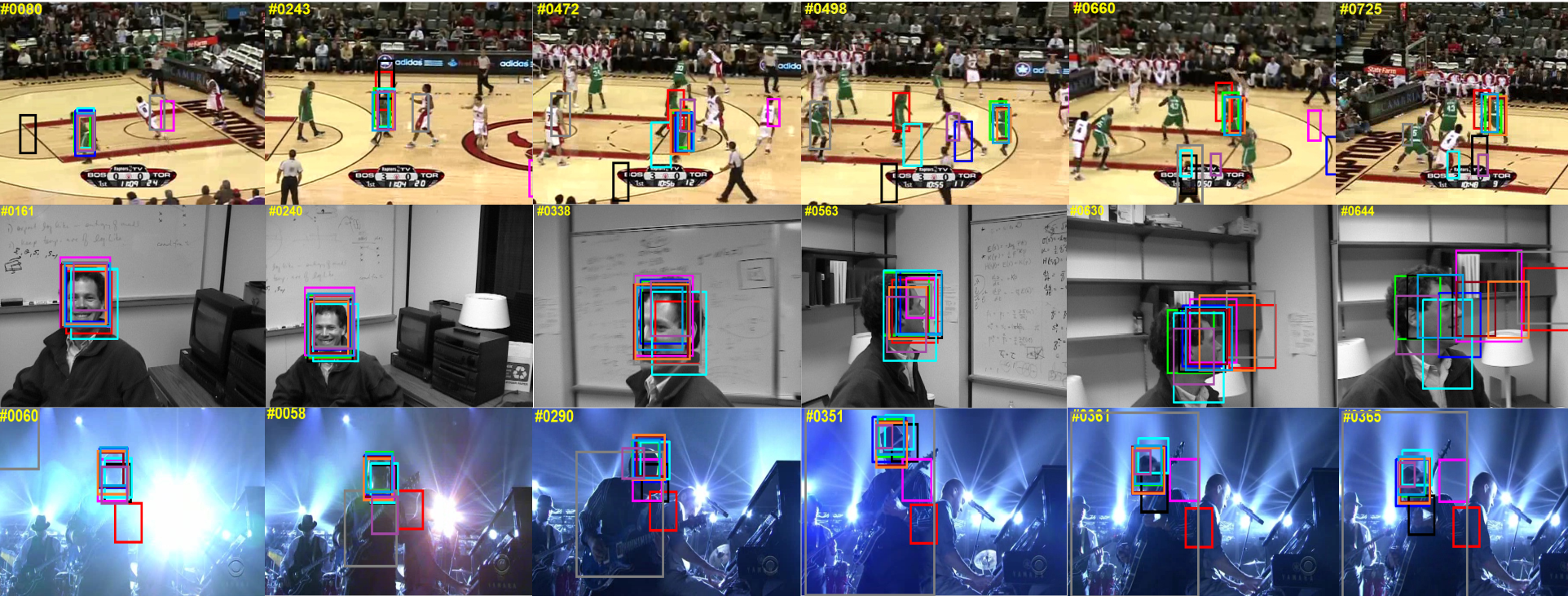}\\
  \includegraphics[width=1\linewidth]{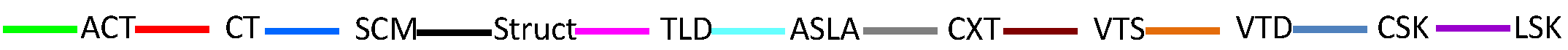}\\
  \caption{Qualitative results of the $11$ trackers over sequences \textit{Basketball}, \textit{Fleetface} and \textit{Shaking} from top to bottom (best viewed on
high-resolution display). Object appearances therein change drastically with deformation.}\label{p1}
\end{figure}

{\bf Deformation:} Figure~\ref{p1} shows the tracking results of three challenging sequences with deformation attributes. In the \textit{Basketball} sequence, the target undergos great changes as the player runs around, especially interferences from other plays. We observe that Struct, CXT, and TLD drift once other players hide the target (e.g., $\#34$). The SCM, ASLA, CT, CSK and LSK drift when the object appearance begins to vary (e.g., $\#472$). VTD, VTS and our ACT method are able to track the target in the whole sequence successfully. Our tracker can deal with deformation well due to its online appearance update and trajectory rectification strategies.

The \textit{Fleetface} sequence suffers from scale changes, appearance varies, and background distraction when the object walks around the room. Many methods fail to track the object when the object turns his head, which results in dramatically appearance changes. Challenges also come from the interference caused by bookshelf, because the color and texture information is similar to object at that time. ASLA, Struct and our ACT methods achieve well performance on this sequence.

In the \textit{Shaking} sequence, the object undergoes both illumination change and pose variation. CSK, SCM, and VTD are able to track the object in this sequence, but with a lower success rate than our method.

\begin{figure}
  \centering
  \includegraphics[width=1\linewidth]{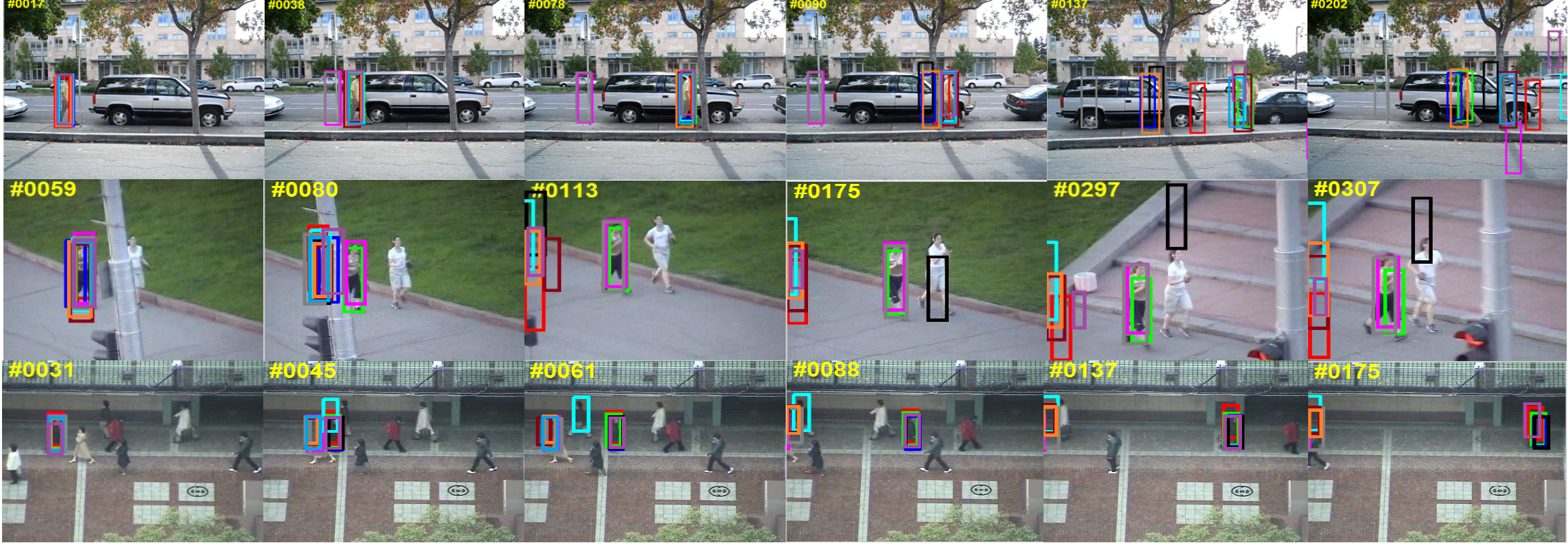}\\
  \includegraphics[width=1\linewidth]{figure/leg.png}\\
  \caption{Qualitative results of the $11$ trackers over sequences
  \textit{David3}, \textit{Jogging} and \textit{Subway} from top to bottom (best viewed on
high-resolution display). Object appearances therein changes drastically with heavy occlusion.}\label{p2}
\end{figure}

{\bf Heavy occlusion:} The targets in the sequences of Figure~\ref{p2} undergo heavy occlusions from other objects. Furthermore, the targets in these sequences suffer from severe pose variations when the pedestrians turn round. Both make these sequences much challenging. Overall, ACT shows favorable performance to tackle these challenges, which attributes to the adaptive appearance model and online template update mechanism. When the confidence score of ACT decreases greatly to zero, the classifier and template stop updating, which can well prevent the tracker from drifting due to adding inaccurate samples.

\begin{figure}
  \centering
  \includegraphics[width=1\linewidth]{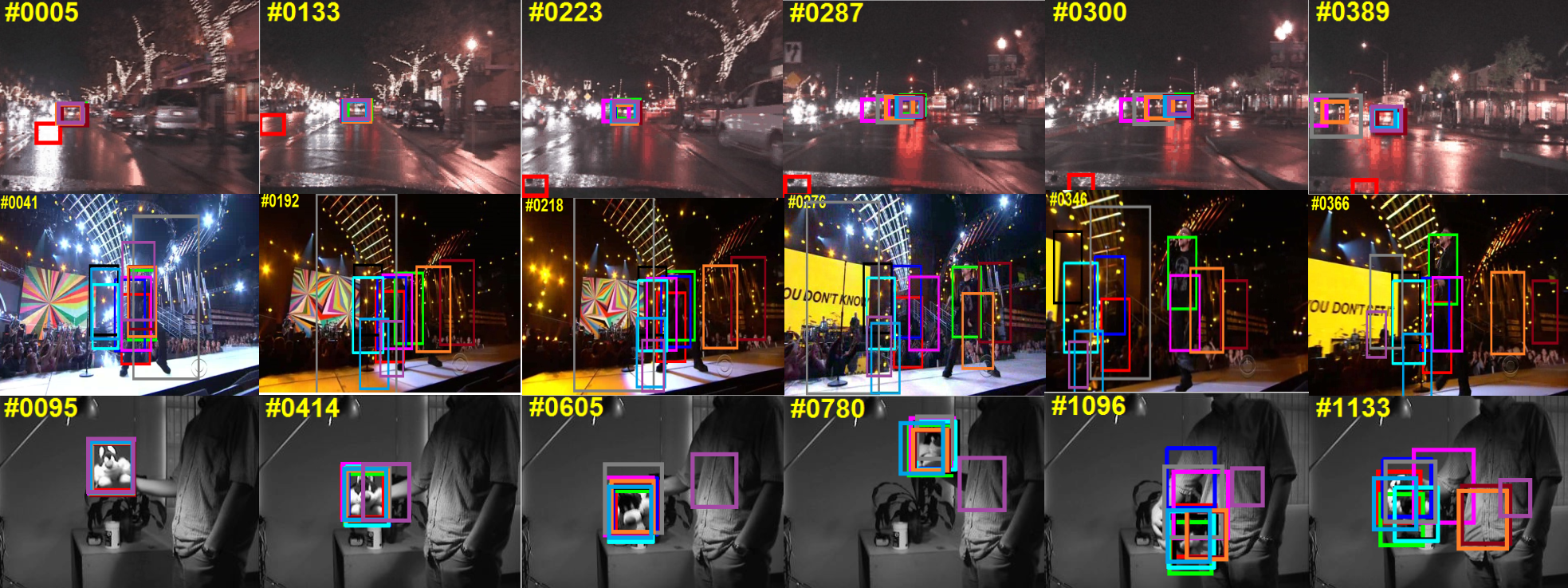}\\
  \includegraphics[width=1\linewidth]{figure/leg.png}\\
  \caption{Qualitative results of the $11$ trackers over sequences
  \textit{CarDark}, \textit{Singer2} and \textit{Sylvester} from top to bottom (best viewed on
high-resolution display). Objects therein undergo illumination changes.}\label{p3}
\end{figure}

{\bf Illumination change:} Figure~\ref{p3} shows the tracking results of three challenging sequences where the targets suffer from drastic illumination changes. In the \textit{CarDark} sequence, a car runs along the street at night that suffers from large changes in environmental illumination and background clutters, and CT, TLD, CXT, and VTD drift gradually (e.g., $\#287$). In contrast, SCM, Struct, ASLA, VTS, CSK, LSK and our ACT achieve much better performance. For the \textit{Singer2} sequence, there is small contrast between the object and background besides illumination changes. Many trackers drift to the background at the beginning of this sequence when stage light changes drastically (e.g., $\#41$). For the \textit{Sylvester} sequence, challenges like IV, OPR, and IPR make it difficult to robustly track. Notwithstanding, our tracker achieve favorable performance due to its adaptive local appearance model.

\begin{figure}
  \centering
  \includegraphics[width=1\linewidth]{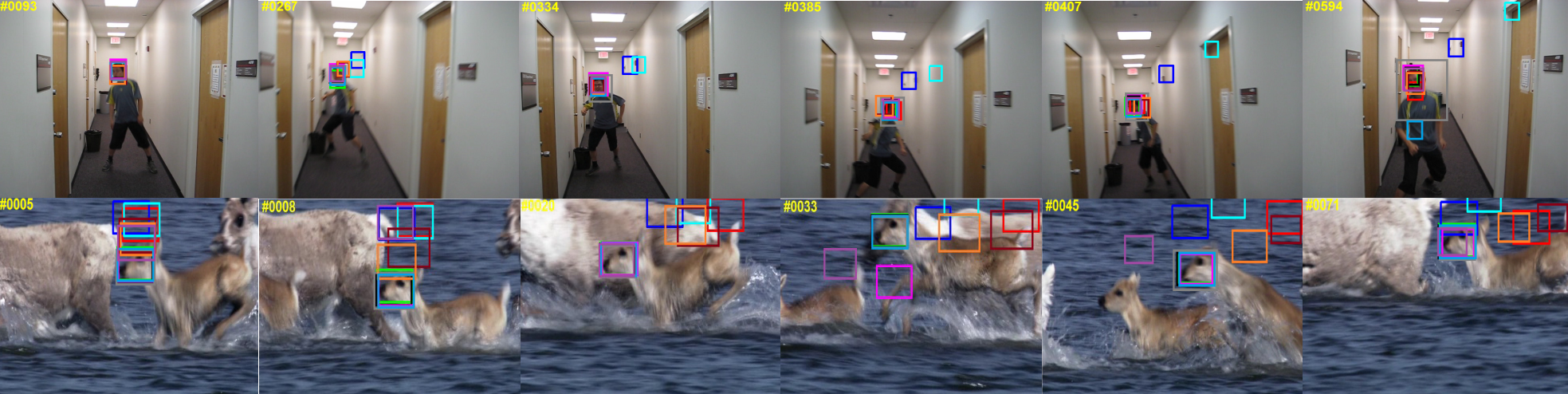}\\
  \includegraphics[width=1\linewidth]{figure/leg.png}\\
  \caption{Qualitative tracking results of the 11 trackers over sequences
  \textit{boy}, \textit{Deer} from top to bottom (best viewed on
high-resolution display. Objects therein undergo other challenges.}\label{p4}
\end{figure}

{\bf Other challenges:} Figure~\ref{p4} presents the tracking results in which many other challenges occur in these sequences, such as MB, FM, BC, SV, etc. In the \textit{Boy} sequence, a boy jumps regularly, causing MB and SV in his face, making it difficult to track. Our ACT performs well in this sequence because of online feature template update. The target in the \textit{Deer} sequence suffers from MB, FM and BC, our ACT works well due to its online trajectory rectification strategy that can prevent the model update from inaccurate samples.

\subsection{Analysis of ACT}
\begin{figure}[t]
 \includegraphics[width=0.45\textwidth]{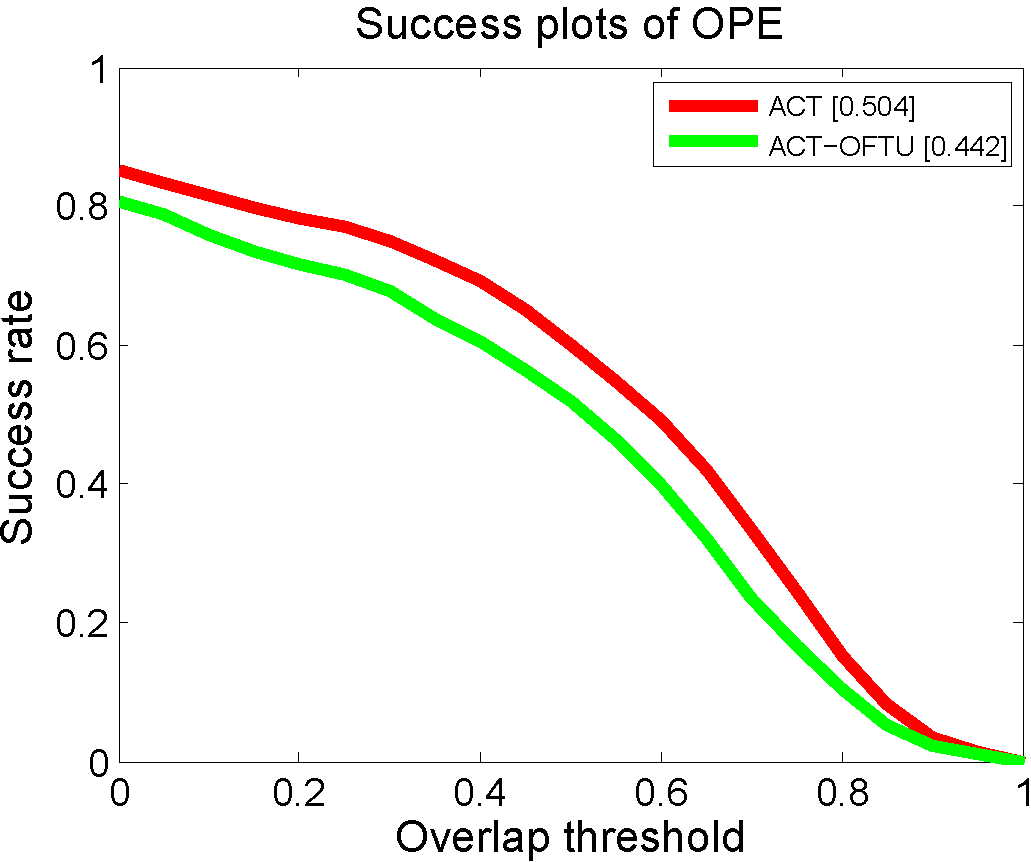}
 \hspace{3ex}
 \includegraphics[width=0.46\textwidth]{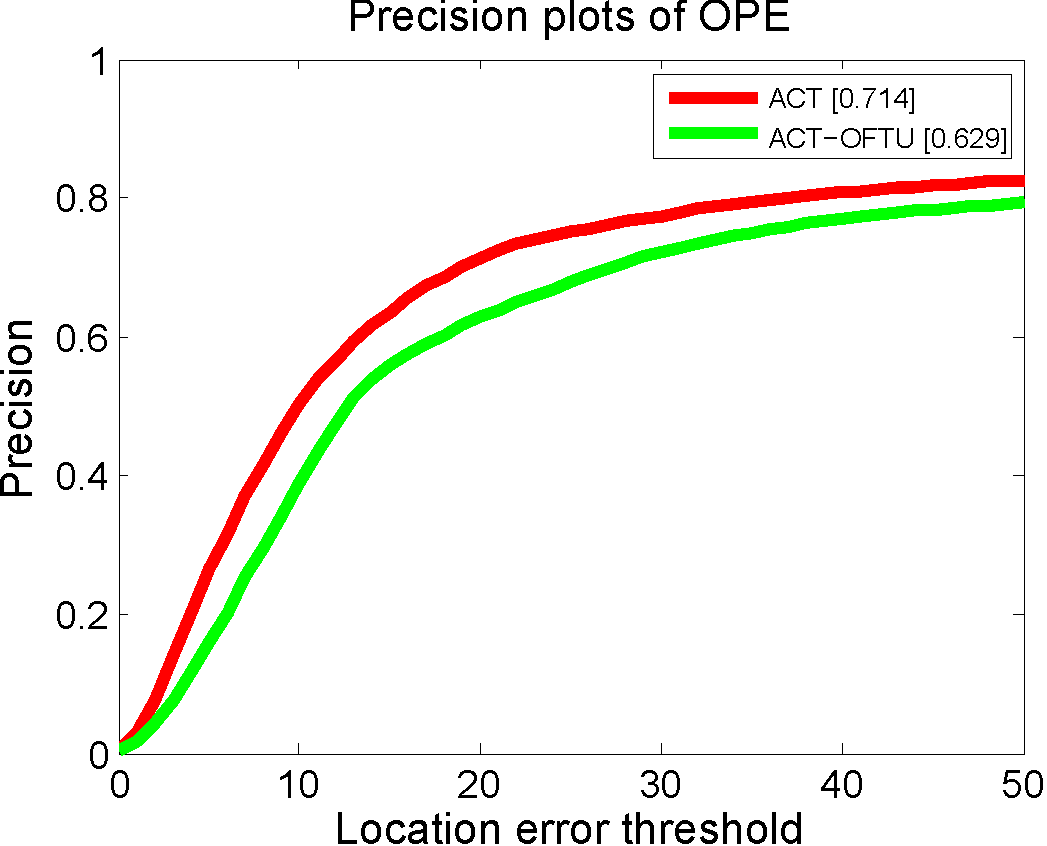}
 \caption{The success plots and the precision plots for ACT and ACT-OFTU}\label{OFTU1}
\end{figure}

{\bf Online feature template update (OFTU):}
To verify the effectiveness of OFTU, we develop a tracker named ACT-OFTU that removes the component of OFTU in ACT. The quantitative results are illustrated in Figure~\ref{OFTU1}, where we can observe that ACT achieves much better performance than ACT-OFTU. OFTU emphasizes the importance of object appearance variance over time, where the stable templates are preserved. Furthermore, the update part in the templates takes into account the appearance variations, which can well adapt the target appearance variations over time.

%

\begin{figure}[t]
 \includegraphics[width=0.45\textwidth]{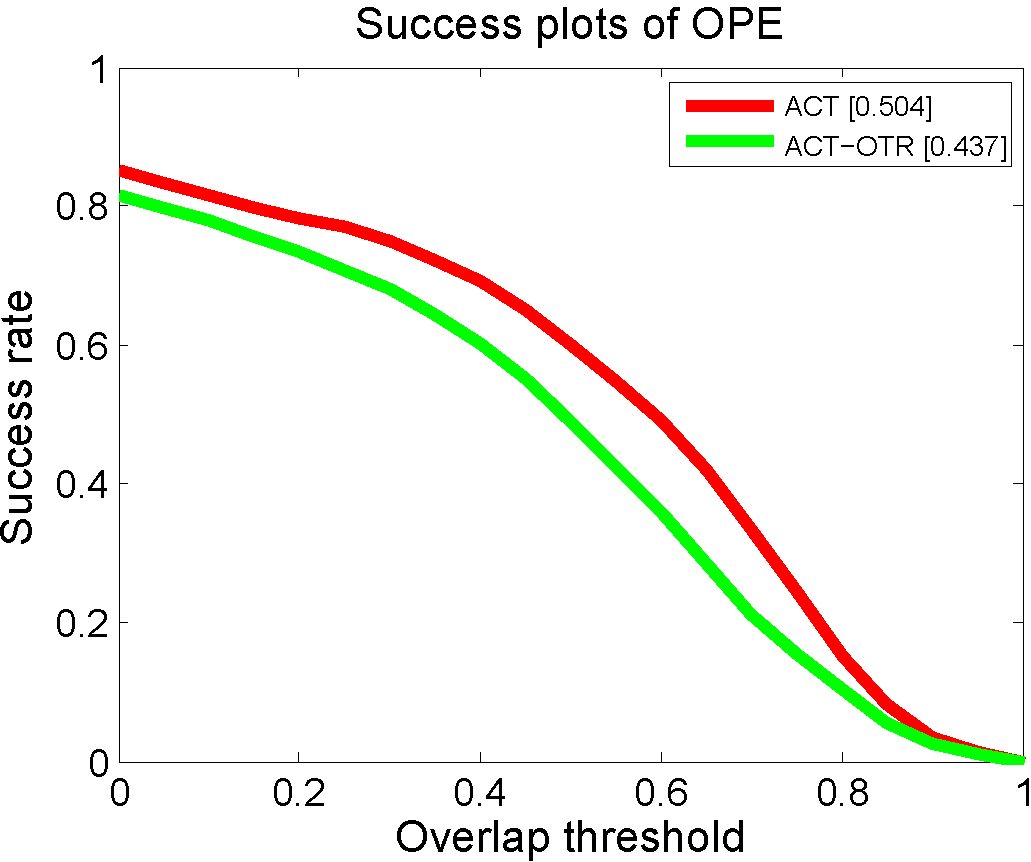}
 \hspace{3ex}
 \includegraphics[width=0.46\textwidth]{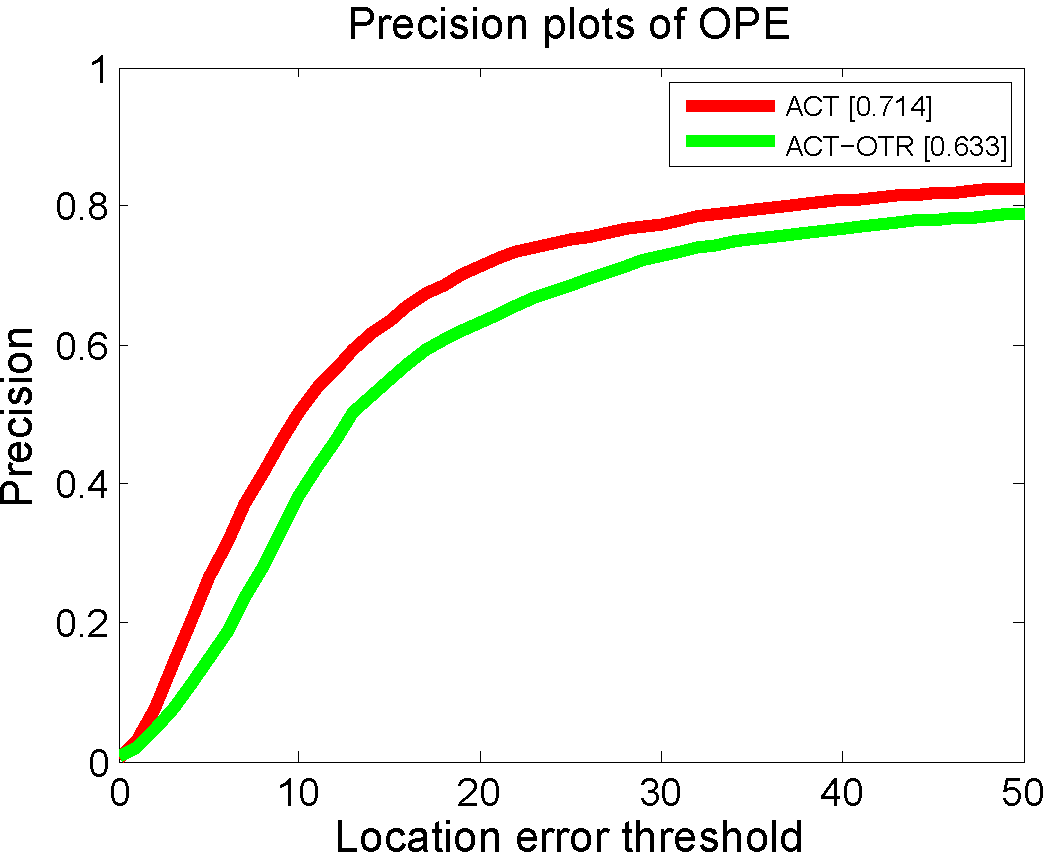}
 \caption{The success plots and the precision plots for ACT and ACT-OTR}\label{OTR1}
\end{figure}

{\bf Online trajectory rectification (OTR):}
We design a tracker called ACT-OTR to justify the effectiveness of OTR in ACT. Figure~\ref{OTR1} illustrates the quantitative results, where ACT outperforms ACT-OTR by a large margin, demonstrating the effectiveness of OTR that can well prevent the model update from inaccurate samples.

\begin{figure}[t]
 \includegraphics[width=0.45\textwidth]{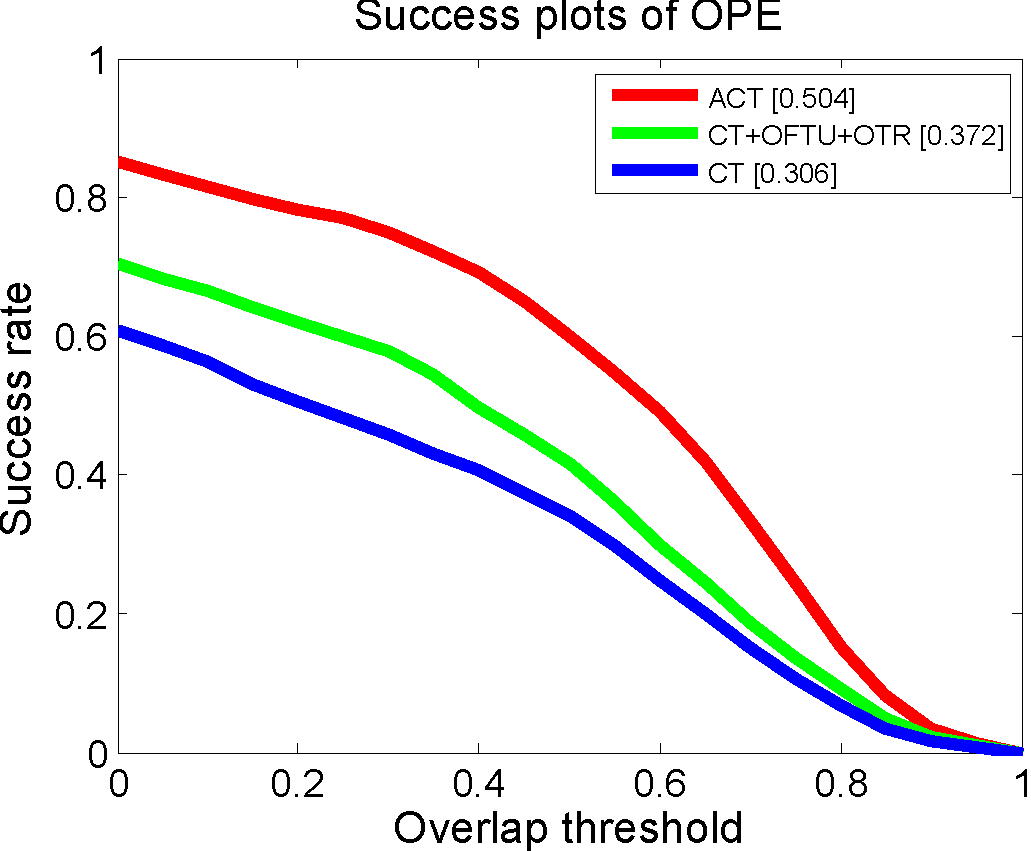}
 \hspace{3ex}
 \includegraphics[width=0.46\textwidth]{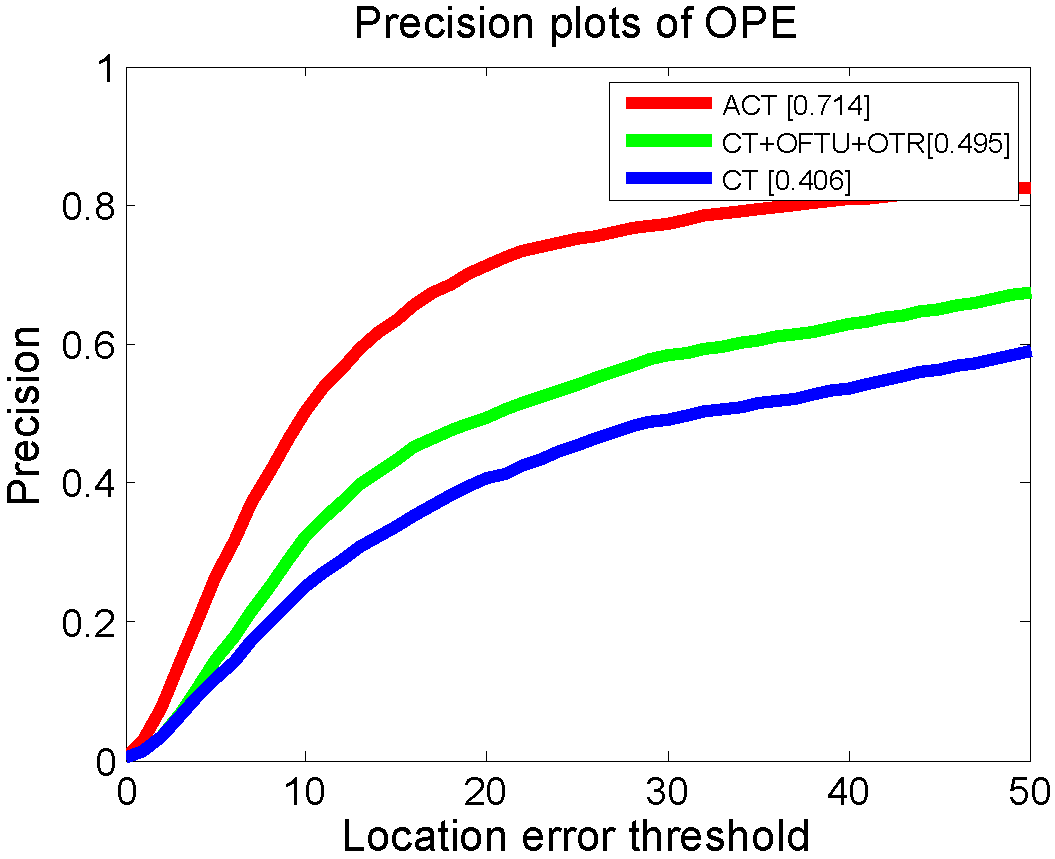}
 \caption{The success plots and the precision plots for CT, CT+OFTU+OTR, and ACT}\label{CT1}
\end{figure}

{\bf Online vector boosting feature selection (OVBFS):}
To justify the effectiveness of OVBFS, we construct a tracker named CT+OFTU+OTR that replaces the OVBFS component in ACT with the compressive Haar-like features in CT~\cite{zhang2014fast}. The quantitative results are shown in Figure~\ref{CT1}, where CT performs unfavorably due to the fact that the feature templates may select noninformative features when they fall into the textureless regions, but with the OFTU and OTR, CT improves its performance significantly, which demonstrates the effectiveness of OFTU and OTR in ACT.
%

%
\section{Conclusions}

In this paper, we have proposed a novel adaptive compressive tracking algorithm that improves the CT algorithm~\cite{zhang2012real} by a significantly large margin on the CVPR2013 tracking benchmark~\cite{wu2013online}. The proposed algorithm mainly includes three components: First, a novel vector boosting feature selection strategy has been proposed to design an effective appearance model; Second, a simple conservative model update strategy has been adopted that can preserve the stable information while filtering out the noisy appearance variations during tracking; Third, a simple and effective trajectory rectification strategy has been developed that can refine the tracking location when possible inaccurate tracking occurs. Extensive evaluations on the CVPR2013 tracking benchmark have demonstrated the superior performance of the proposed algorithm over other state-of-the-art tracking algorithms.

\bibliographystyle{model1-num-names}
\bibliography{egbib}







\end{document}